# Subspace Clustering by Block Diagonal Representation

Canyi Lu, *Student Member, IEEE*, Jiashi Feng, Zhouchen Lin, *Senior Member, IEEE*, Tao Mei, *Senior Member, IEEE*, and Shuicheng Yan, *Fellow, IEEE*

**Abstract**—This paper studies the subspace clustering problem. Given some data points approximately drawn from a union of subspaces, the goal is to group these data points into their underlying subspaces. Many subspace clustering methods have been proposed and among which sparse subspace clustering and low-rank representation are two representative ones. Despite the different motivations, we observe that many existing methods own the common block diagonal property, which possibly leads to correct clustering, yet with their proofs given case by case. In this work, we consider a general formulation and provide a unified theoretical guarantee of the block diagonal property. The block diagonal property of many existing methods falls into our special case. Second, we observe that many existing methods approximate the block diagonal representation matrix by using different structure priors, e.g., sparsity and low-rankness, which are indirect. We propose the first block diagonal matrix induced regularizer for directly pursuing the block diagonal matrix. With this regularizer, we solve the subspace clustering problem by Block Diagonal Representation (BDR), which uses the block diagonal structure prior. The BDR model is nonconvex and we propose an alternating minimization solver and prove its convergence. Experiments on real datasets demonstrate the effectiveness of BDR.

✦

## 1 INTRODUCTION

As we embark on the big data era – in which the amount of the generated and collected data increases quickly, the data processing and understanding become impossible in the raw form. Looking for the compact representation of data by exploiting the structure of data is crucial in understanding the data with minimal storage. It is now widely known that many high dimensional data can be modeled as samples drawn from the union of multiple low-dimensional linear subspaces. For example, motion trajectories in a video [8], face images [2], hand-written digits [13] and movie ratings [45] can be approximately represented by subspaces, with each subspace corresponding to a class or category. Such a subspace structure has been very widely used for the data processing and understanding in supervised learning, semi-supervised learning and many other tasks [5], [6], [44], [46]. In this work, we are interested in the task of *subspace clustering*, whose goal is to group (or cluster) the data points which approximately lie in linear subspaces into clusters with each cluster corresponding to a subspace. Subspace clustering has many applications in computer vision [16], [28], e.g., motion segmentation, face clustering and image segmentation, hybrid system identification in control [1], community clustering in social networks [15], to name a few. Note that subspace clustering is a data clustering task but with the additional assumption that the sampled data have the approximately linear subspace structure. Such data points are not necessarily locally distributed. The traditional clustering methods, e.g., spectral clustering [31], which use the spatial proximity of the data in each cluster are not applicable to subspace clustering. We need some more advanced methods for subspace clustering by utilizing the subspace structure as a prior.

**Notations.** We denote matrices by boldface capital letters, e.g., $\mathbf{A}$, vectors by boldface lowercase letters, e.g., $\mathbf{a}$, and scalars by lowercase letters, e.g., $a$. We denote $a_{ij}$ or $\mathbf{A}_{ij}$ as the $(i,j)$-th entry of $\mathbf{A}$. The matrix columns and rows are denoted by using $[\cdot]$ with subscripts, e.g., $[\mathbf{A}]_{i,:}$ is the $i$-th row, and $[\mathbf{A}]_{:,j}$ is the $j$-th column. The absolute matrix of $\mathbf{A}$, denoted by $|\mathbf{A}|$, is the absolute value of the elements of $\mathbf{A}$. We denote $\text{diag}(\mathbf{A})$ as a vector with its $i$-th element being the $i$-th diagonal element of $\mathbf{A} \in \mathbb{R}^{n \times n}$, and $\text{Diag}(\mathbf{a})$ as a diagonal matrix with its $i$-th element on the diagonal being $a_i$. The all one vector is denoted as $\mathbf{1}$. The identity matrix is denoted as $\mathbf{I}$. If $\mathbf{A}$ is positive semi-definite, we denote $\mathbf{A} \succeq 0$. For symmetric matrices $\mathbf{A}, \mathbf{B} \in \mathbb{R}^{n \times n}$, we denote $\mathbf{A} \preceq \mathbf{B}$ or $\mathbf{B} \succeq \mathbf{A}$ if $\mathbf{B} - \mathbf{A} \succeq 0$. If all the elements of $\mathbf{A}$ are nonnegative, we denote $\mathbf{A} \geq 0$. The trace of a square matrix $\mathbf{A}$ is denoted as $\text{Tr}(\mathbf{A})$. We define $[\mathbf{A}]_+ = \max(0, \mathbf{A})$ which gives the nonnegative part of the matrix.

Some norms will be used, e.g., $\ell_0$-norm $\|\mathbf{A}\|_0$ (number of nonzero elements), $\ell_1$-norm $\|\mathbf{A}\|_1 = \sum_{ij} |a_{ij}|$, Frobenius norm (or $\ell_2$-norm of a vector) $\|\mathbf{A}\| = \sqrt{\sum_{ij} a_{ij}^2}$, $\ell_{2,1}$-norm $\|\mathbf{A}\|_{2,1} = \sum_j \|[\mathbf{A}]_{:,j}\|$, $\ell_{1,2}$-norm $\|\mathbf{A}\|_{1,2} = \sum_i \|[\mathbf{A}]_{i,:}\|$, spectral norm $\|\mathbf{A}\|_2$ (largest singular value), $\ell_\infty$-norm $\|\mathbf{A}\|_\infty = \max_{ij} |a_{ij}|$ and nuclear norm $\|\mathbf{A}\|_*$ (sum of all singular values).

---

- C. Lu, J. Feng and S. Yan are with the Department of Electrical and Computer Engineering, National University of Singapore, Singapore (e-mail: canyilu@gmail.com; elefjia@nus.edu.sg; eleyans@nus.edu.sg).
- Z. Lin is with the Key Laboratory of Machine Perception (MOE), School of EECS, Peking University, Beijing 100871, China, and also with the Cooperative Medianet Innovation Center, Shanghai Jiaotong University, Shanghai 200240, China (email: zlin@pku.edu.cn).
- T. Mei is with the Microsoft Research Asia (e-mail: tmei@microsoft.com).

## 1.1 Related Work

Due to the numerous applications in computer vision and image processing, during the past two decades, subspace clustering has been extensively studied and many algorithms have been proposed to tackle this problem. According to their mechanisms of representing the subspaces, existing works can be roughly divided into four main categories: mixture of Gaussian, matrix factorization, algebraic, and spectral-type methods. The mixture of Gaussian based methods model the data points as independent samples drawn from a mixture of Gaussian distributions. So subspace clustering is converted to the model estimation problem and the estimation can be performed by using the Expectation Maximization (EM) algorithm. Representative methods are K-plane [3] and Q-flat [37]. The limitations are that they are sensitive to errors and the initialization due to the optimization mechanism. The matrix factorization based methods, e.g., [8], [12], tend to reveal the data segmentation based on the factorization of the given data matrix. They are sensitive to data noise and outliers. Generalized Principal Component Analysis (GPCA) [38] is a representative algebraic method for subspace clustering. It fits the data points with a polynomial. However, this is generally difficult due to the data noise and its cost is high especially for high-dimensional data. Due to the simplicity and outstanding performance, the spectral-type methods attract more attention in recent years. We give a more detailed review of this type of methods as follows.

The spectral-type methods use the spectral clustering algorithm [31] as the framework. They first learn an affinity matrix to find the low-dimensional embedding of data and then k-means is applied to achieve the final clustering result. The main difference among different spectral-type methods lies in the different ways of affinity matrix construction. The entries of the affinity matrix (or graph) measure the similarities of the data point pairs. Ideally, if the affinity matrix is block diagonal, i.e., the between-cluster affinities are all zeros, one may achieve perfect data clustering by using spectral clustering. The way of affinity matrix construction by using the typical Gaussian kernel, or other local information based methods, e.g., Local Subspace Affinity (LSA) [43], may not be a good choice for subspace clustering since the data points in a union of subspaces may be distributed arbitrarily but not necessarily locally. Instead, a large body of affinity matrix construction methods for subspace clustering by using global information have been proposed in recent years, e.g., [10], [20], [23], [25], [26], [27], [40]. The main difference among them lies in the used regularization for learning the representation coefficient matrix.

Assume that we are given the data matrix $\mathbf{X} \in \mathbb{R}^{D \times n}$, where each column of $\mathbf{X}$ belongs to a union of $k$ subspaces $\{\mathcal{S}\}_{i=1}^{k}$. Each subspace $i$ contains $n_i$ data samples with $\sum_{i=1}^{k} n_i = n$. Let $\mathbf{X}_i \in \mathbb{R}^{D \times n_i}$ denote the submatrix in $\mathbf{X}$ that belongs to $\mathcal{S}_i$. Without loss of generality, let $\mathbf{X} = [\mathbf{X}_1, \mathbf{X}_2, \cdots, \mathbf{X}_k]$ be ordered according to their subspace membership. We discuss the case that the sampled data are noise free. By taking advantage of the subspace structure, the sampled data points obey the so called self-expressiveness property, i.e., each data point in a union of subspaces can be well represented by a linear combination of other points in the dataset. This can be formulated as

$$\mathbf{X} = \mathbf{X}\mathbf{Z}, \tag{1}$$

where $\mathbf{Z} \in \mathbb{R}^{n \times n}$ is the representation coefficient matrix. The choice of $\mathbf{Z}$ is usually not unique and the goal is to find certain $\mathbf{Z}$ such that it is discriminative for subspace clustering. In the ideal case, we are looking for a linear representation $\mathbf{Z}$ such that each sample is represented as a linear combination of samples belonging to the same subspace, i.e., $\mathbf{X}_i = \mathbf{X}_i \mathbf{Z}_i$, where $\mathbf{Z}_i$ is expected not to be an identity matrix. In this case, $\mathbf{Z}$ in (1) has the $k$-block diagonal structure[1], i.e.,

$$\mathbf{Z} = \begin{bmatrix} \mathbf{Z}_1 & 0 & \cdots & 0 \\ 0 & \mathbf{Z}_2 & \cdots & 0 \\ \vdots & \vdots & \ddots & \vdots \\ 0 & 0 & \cdots & \mathbf{Z}_k \end{bmatrix}, \ \mathbf{Z}_i \in \mathbb{R}^{n_i \times n_i}. \tag{2}$$

So the above $\mathbf{Z}$ reveals the true membership of data $\mathbf{X}$. If we apply spectral clustering on the affinity matrix defined as $(|\mathbf{Z}| + |\mathbf{Z}^\top|)/2$, then we may get correct clustering. So the block diagonal matrix plays a central role in the analysis of subspace clustering, though there has no "ground-truth" $\mathbf{Z}$ (or it is not necessary). We formally give the following definition.

**Definition 1. (Block Diagonal Property (BDP))** *Given the data matrix* $\mathbf{X} = [\mathbf{X}_1, \mathbf{X}_2, \cdots, \mathbf{X}_k]$ *drawn from a union of* $k$ *subspaces* $\{\mathcal{S}_i\}_{i=1}^{k}$, *we say that* $\mathbf{Z}$ *obeys the Block Diagonal Property Property if* $\mathbf{Z}$ *is* $k$-*block diagonal as in (2), where the nonzero entries* $\mathbf{Z}_i$ *correspond to only* $\mathbf{X}_i$.

Note that the concepts of the $k$-block diagonal matrix and block diagonal property have some connections and differences. The block diagonal property is specific for subspace clustering problem but $k$-block diagonal matrix is not. A matrix obeying the block diagonal property is $k$-block diagonal, but not vice versa. The block diagonal property further requires that each block corresponds one-to-one with each subject of data.

Problem (1) may have many feasible solutions and thus the regularization is necessary to produce the block diagonal solution. Motivated by the observation that the block diagonal solution in (2) is sparse, the Sparse Subspace Clustering (SSC) [10] finds a sparse $\mathbf{Z}$ by $\ell_0$-norm minimizing. However, this leads to an NP-hard problem and the $\ell_1$-norm is used as the convex surrogate of $\ell_0$-norm. This leads to the following convex program

$$\min_{\mathbf{Z}} \ \|\mathbf{Z}\|_1, \ \text{s.t.} \ \mathbf{X} = \mathbf{X}\mathbf{Z}, \text{diag}(\mathbf{Z}) = 0. \tag{3}$$

It is proved that the optimal solution $\mathbf{Z}$ by SSC satisfies the block diagonal property when the subspaces are independent.

**Definition 2. (Independent subspaces)** *A collection of subspaces* $\{\mathcal{S}_i\}_{i=1}^{k}$ *is said to be independent if* $dim(\oplus_{i=1}^{n} \mathcal{S}_i) = \sum_{i=1}^{n} dim(\mathcal{S}_i)$, *where* $\oplus$ *denotes the direct sum operator.*

---

1. In this work, we say that a matrix is $k$-block diagonal if it has *at least* $k$ connected components (blocks). The block diagonality is up to a permutation, i.e., if $\mathbf{Z}$ is $k$-block diagonal, then $\mathbf{P}^\top \mathbf{Z} \mathbf{P}$ is still $k$-block diagonal for any permutation matrix $\mathbf{P}$. See also the discussions in Section 3.1.

TABLE 1: A summary of existing spectral-type subspace clustering methods based on different choices of $f$ and $\Omega$.

| Methods | $f(\mathbf{Z}, \mathbf{X})$ | $\Omega$ |
|---|---|---|
| SSC [10] | $\|\mathbf{Z}\|_1$ | $\{\mathbf{Z}|\mathrm{diag}(\mathbf{Z})=0\}$ |
| LRR [20] | $\|\mathbf{Z}\|_*$ | - |
| MSR [27] | $\|\mathbf{Z}\|_1 + \lambda\|\mathbf{Z}\|_*$ | $\{\mathbf{Z}|\mathrm{diag}(\mathbf{Z})=0\}$ |
| SSQP [40] | $\|\mathbf{Z}^\top\mathbf{Z}\|_1$ | $\{\mathbf{Z}|\mathrm{diag}(\mathbf{Z})=0, \\ \mathbf{Z}\geq 0\}$ |
| LSR [26] | $\|\mathbf{Z}\|^2$ | - |
| CASS [23] | $\sum_j \|\mathbf{X}\mathrm{Diag}([\mathbf{Z}]_{:,j})\|_*$ | $\{\mathbf{Z}|\mathrm{diag}(\mathbf{Z})=0\}$ |

[1] $\Omega$ is not specified if there has no restriction on $\mathbf{Z}$.

Another important spectral-type method is Low-Rank Representation (LRR) [20]. It seeks a low-rank coefficient matrix by nuclear norm minimization

$$\min_{\mathbf{Z}} \ \|\mathbf{Z}\|_*, \ \text{s.t. } \mathbf{X} = \mathbf{X}\mathbf{Z}. \tag{4}$$

The above problem has a unique closed form solution $\mathbf{Z} = \mathbf{V}\mathbf{V}^\top$, where $\mathbf{V}$ is from the skinny SVD of $\mathbf{X} = \mathbf{U}\mathbf{S}\mathbf{V}^\top$. This matrix, termed Shape Interaction Matrix (SIM) [8], has been widely used for subspace segmentation. It also enjoys the block diagonal property when the subspaces are independent [20].

Beyond SSC and LRR, many other subspace clustering methods, e.g., [23], [26], [27], [40], have been proposed and they all fall into the following formulation

$$\min f(\mathbf{Z}, \mathbf{X}), \ \text{s.t. } \mathbf{X} = \mathbf{X}\mathbf{Z}, \mathbf{Z} \in \Omega, \tag{5}$$

where $\Omega$ is some matrix set. The main difference lies in the choice of the regularizer or objective. For example, the Multi-Subspace Representation (MSR) [27] combines the idea of SSC and LRR, while the Least Squares Regression (LSR) [26] simply uses $\|\mathbf{Z}\|^2$ and it is efficient due to a closed form solution. See Table 1 for a summary of existing spectral-type methods. An important common property for the methods in Table 1 is that their solutions all obey the block diagonal property under certain subspace assumption (all require independent subspaces assumption except SSQP [40] that requires orthogonal subspaces assumption). Their proofs use specific properties of their objectives.

Beyond the independent subspaces assumption, some other subspaces assumptions are proposed to analyze the block diagonal property in different settings [10], [34], [35], [41], [42]. However, the block diagonal property of $\mathbf{Z}$ does not guarantee the correct clustering, since each block may not be fully connected. For example, the work [10] shows that the block diagonal property holds for SSC when the subspaces are disjoint and the angles between subspace pairs are large enough. Such an assumption is weaker than the independent subspaces assumption, but the price is that SSC suffers from the so-called "graph connectivity" issue [30]. This issue is also related to the correlation of the columns of the data matrix [26]. As will be seen in Theorem 3 given later, the $\ell_1$-minimization in SSC makes not only the between-cluster connections sparse, but also the inner-cluster connections sparse. In this case, the clustering results obtained by spectral clustering may not be correct. Nevertheless, the block diagonal property is the condition that verifies the design intuition of the spectral-type methods. If the obtained coefficient matrix $\mathbf{Z}$ obeys the block diagonal property and each block is fully connected ($\mathbf{Z}$ is not "too sparse"), then we immediately get the correct clustering.

The block diagonal property of the solutions by different methods in Table 1 is common under certain subspace assumptions. However, in real applications, due to the data noise or corruptions, the required assumptions usually do not hold and thus the block diagonal property is violated. By taking advantage of the $k$-block diagonal structure as a prior, the work [11] considers SSC and LRR with an additional hard Laplacian constraint, which enforces $\mathbf{Z}$ to be $k$-block diagonal with exact $k$ connected blocks. Though such a $k$-block diagonal solution may not obey the block diagonal property without additional subspace assumption, it is verified to be effective in improving the clustering performance of SSC and LRR in some applications. Due to the nonconvexity, this model suffers from some issues: the used stochastic sub-gradient descent solver may not be stable; and the theoretical convergence guarantee is relatively weak due to the required assumptions on the data matrix.

## 1.2 Contributions

In this work, we focus on the most recent spectral-type subspace clustering methods due to their simplicity and effectiveness. From the above review, it can be seen that the key difference between different spectral-type subspace clustering methods (as given in Table 1) is the used regularizer on the representation matrix $\mathbf{Z}$. Their motivations for the design intuition may be quite different, but all have the common property that their solutions obey the block diagonal property under certain subspace assumption. However, their proofs of such a property are given case by case by using specific properties of the models. Moreover, existing methods in Table 1 are indirect as their regularizers are not induced by the block diagonal matrix structure. The method in [11] that enforces the solution to be $k$-block diagonal with exact $k$ connected blocks by a hard constraint is a direct method. But such a constraint may be too restrictive since the $k$-block diagonal matrix is not necessary for correct clustering when using spectral clustering. A soft regularizer instead of the hard constraint may be more flexible. Motivated by these observations, we raise several interesting questions:

1. Consider the general model (5), what kind of objective $f$ guarantees that the solutions obey the block diagonal property?
2. Is it possible to give a unified proof of the block diagonal property by using common properties of the objective $f$?
3. How to design a soft block diagonal regularizer which encourages a matrix to be or close to be $k$-block diagonal? When applying it to subspace clustering, how to solve the block diagonal regularized problem efficiently with the convergence guarantee?

We aim to address the above questions and in particular we make the following contributions[2]:

1. We propose the Enforced Block Diagonal (EBD) conditions and prove in a unified manner that if the objective function in (5) satisfies the EBD conditions, the solutions

2. Part of this work is extended from our conference paper [26].

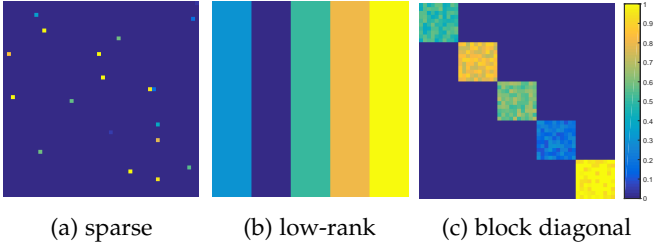

(a) sparse     (b) low-rank     (c) block diagonal

Fig. 1: Illustrations of three interesting structures of matrix: sparse, low-rank and block diagonal matrices. The first two are extensively studied before. This work focuses on the pursuit of block diagonal matrix.

to (5) obey the block diagonal property when the subspaces are independent. We show that the EBD conditions are not restrictive and a large family of norms and their combinations satisfy these conditions. The block diagonal property of existing methods in Table 1 falls into our special case.

2. We propose a *k-block diagonal regularizer* which encourages a nonnegative symmetric matrix to be $k$-block diagonal. Beyond the sparsity and low-rankness, we would like to emphasize that the block diagonal matrix is another interesting structure and our proposed block diagonal regularizer is the first soft regularizer for pursuing such a structure. The regularizer plays a similar role as the $\ell_0$- or $\ell_1$-norm for pursuing sparsity and the rank function or nuclear norm for pursuing low-rankness. See Figure 1 for intuitive illustrations of the three structured matrices.

3. We propose the Block Diagonal Representation (BDR) method for subspace clustering by using the block diagonal regularizer. Compared with the regularizers used in existing methods, BDR is more direct as it uses the block diagonal structure prior. A disadvantage of the BDR model is that it is nonconvex due to the block diagonal regularizer. We solve it by an alternating minimization method and prove the convergence without restrictive assumptions. Experimental analysis on several real datasets demonstrates the effectiveness of our approach.

## 2 THEORY OF BLOCK DIAGONAL PROPERTY

In this section, considering problem (5), we develop the unified theory for pursuing solutions which obey the block diagonal property. We first give an important property of the feasible solution to (5). This will lead to our EBD conditions.

**Theorem 1.** *Consider a collection of data points drawn from $k$ independent subspaces $\{S_i\}_{i=1}^k$ of dimensions $\{d_i\}_{i=1}^k$. Let $\mathbf{X} = [\mathbf{X}_1, \cdots, \mathbf{X}_k] \in \mathbb{R}^{D \times n}$, where $\mathbf{X}_i \in \mathbb{R}^{D \times n_i}$ denotes the data point drawn from $S_i$, $\text{rank}(\mathbf{X}_i) = d_i$ and $\sum_{i=1}^k n_i = n$. For any feasible solution $\mathbf{Z}^* \in \mathbb{R}^{n \times n}$ to the following system*

$$\mathbf{X} = \mathbf{X}\mathbf{Z}, \qquad (6)$$

*decompose it into two parts, i.e., $\mathbf{Z}^* = \mathbf{Z}^B + \mathbf{Z}^C$, where*

$$\mathbf{Z}^B = \begin{bmatrix} \mathbf{Z}_1^* & 0 & \cdots & 0 \\ 0 & \mathbf{Z}_2^* & \cdots & 0 \\ \vdots & \vdots & \ddots & \vdots \\ 0 & 0 & \cdots & \mathbf{Z}_k^* \end{bmatrix}, \ \mathbf{Z}^C = \begin{bmatrix} 0 & * & \cdots & * \\ * & 0 & \cdots & * \\ \vdots & \vdots & \ddots & \vdots \\ * & * & \cdots & 0 \end{bmatrix}, \ (7)$$

*with $\mathbf{Z}_i^* \in \mathbb{R}^{n_i \times n_i}$ corresponding to $\mathbf{X}_i$. Then, we have $\mathbf{X}\mathbf{Z}^B = \mathbf{X}$, or equivalently $\mathbf{X}_i \mathbf{Z}_i^* = \mathbf{X}_i$, $i = 1, \cdots, k$, and $\mathbf{X}\mathbf{Z}^C = 0$.*

*Proof.* For any feasible solution $\mathbf{Z}^*$ to problem (6), we assume that $[\mathbf{X}]_{:,j} = [\mathbf{X}\mathbf{Z}^*]_{:,j} \in S_l$ for some $l$. Then $[\mathbf{X}\mathbf{Z}^B]_{:,j} = [\mathbf{X}_1\mathbf{Z}_1, \cdots, \mathbf{X}_k\mathbf{Z}_k]_{:,j} \in S_l$ and $[\mathbf{X}\mathbf{Z}^C]_{:,j} \in \oplus_{i \neq l} S_i$. On the other hand, $[\mathbf{X}\mathbf{Z}^C]_{:,j} = [\mathbf{X}\mathbf{Z}^*]_{:,j} - [\mathbf{X}\mathbf{Z}^B]_{:,j} \in S_l$. This implies that $[\mathbf{X}\mathbf{Z}^C]_{:,j} \in S_l \cap \oplus_{i \neq l} S_i$. By the assumption that the subspaces are independent, we have $S_l \cap \oplus_{i \neq l} S_i = \{0\}$. Thus, $[\mathbf{X}\mathbf{Z}^C]_{:,j} = 0$. Consider the above procedure for all $j = 1, \cdots, n$, we have $\mathbf{X}\mathbf{Z}^C = 0$ and thus $\mathbf{X}\mathbf{Z}^B = \mathbf{X} - \mathbf{X}\mathbf{Z}^C = \mathbf{X}$. The proof is completed. $\square$

Theorem 1 gives the property of the representation matrix $\mathbf{Z}^*$ under the independent subspaces assumption. The result shows that, to represent a data point $[\mathbf{X}]_{:,j}$ in $S_l$, only the data points $\mathbf{X}_l$ from the same subspace $S_l$ have the real contributions, i.e., $\mathbf{X} = \mathbf{X}\mathbf{Z}^B$, while the total contribution of all the data points from other subspaces $\oplus_{i \neq l} S_i$ is zero, i.e., $\mathbf{X}\mathbf{Z}^C = 0$. So Theorem 1 characterizes the underlying representation contributions of all data points. However, such contributions are not explicitly reflected by the representation matrix $\mathbf{Z}^*$ since the decomposition $\mathbf{Z}^* = \mathbf{Z}^B + \mathbf{Z}^C$ is unknown when $\mathbf{Z}^C \neq 0$. In this case, the solution $\mathbf{Z}^*$ to (6) does not necessarily obey the block diagonal property, and thus it does not imply the true clustering membership of data. To address this issue, it is natural to consider some regularization on the feasible solution set of (6) to make sure that $\mathbf{Z}^C = 0$. Then $\mathbf{Z}^* = \mathbf{Z}^B$ obeys the block diagonal property. Previous works show that many regularizers, e.g., the $\ell_1$-norm and many others shown in Table 1, can achieve this end. Now the questions is, what kind of functions leads to a similar effect? Motivated by Theorem 1, we give a family of such functions as below.

**Definition 3.** *(Enforced Block Diagonal (EBD) conditions)* *Given any function $f(\mathbf{Z}, \mathbf{X})$ defined on $(\Omega, \Delta)$, where $\Omega$ is a set consisting of some square matrices and $\Delta$ is a set consisting of matrices with nonzero columns. For any $\mathbf{Z} = \begin{bmatrix} \mathbf{Z}_1 & \mathbf{Z}_3 \\ \mathbf{Z}_4 & \mathbf{Z}_2 \end{bmatrix} \in \Omega$, $\mathbf{Z} \neq 0$, $\mathbf{Z}_1$, $\mathbf{Z}_2 \in \Omega$, and $\mathbf{X} = [\mathbf{X}_1, \mathbf{X}_2]$, where $\mathbf{X}_1$ and $\mathbf{X}_2$ correspond to $\mathbf{Z}_1$ and $\mathbf{Z}_2$, respectively. Let $\mathbf{Z}^B = \begin{bmatrix} \mathbf{Z}_1 & 0 \\ 0 & \mathbf{Z}_2 \end{bmatrix} \in \Omega$. Assume that all the matrices are of compatible dimensions. The EBD conditions for $f$ are*

*(1) $f(\mathbf{Z}, \mathbf{X}) = f(\mathbf{P}^\top \mathbf{Z} \mathbf{P}, \mathbf{X} \mathbf{P})$, for any permutation matrix $\mathbf{P}$, $\mathbf{P}^\top \mathbf{Z} \mathbf{P} \in \Omega$.*

*(2) $f(\mathbf{Z}, \mathbf{X}) \geqslant f(\mathbf{Z}^B, \mathbf{X})$, where the equality holds if and only if $\mathbf{Z} = \mathbf{Z}^B$ (or $\mathbf{Z}_3 = \mathbf{Z}_4 = 0$).*

*(3) $f(\mathbf{Z}^B, \mathbf{X}) = f(\mathbf{Z}_1, \mathbf{X}_1) + f(\mathbf{Z}_2, \mathbf{X}_2)$.*

We have the following remarks for the EBD conditions:

1. The EBD condition (1) is a basic requirement for subspace clustering. It guarantees that the clustering result is invariant to any permutation of the columns of the input data matrix $\mathbf{X}$. Though we assume that $\mathbf{X} = [\mathbf{X}_1, \mathbf{X}_2, \cdots, \mathbf{X}_k]$ is ordered according to the true membership for the simplicity of discussion, the input matrix in problem (5) can be $\tilde{\mathbf{X}} = \mathbf{X}\mathbf{P}$, where $\mathbf{P}$ can be any permutation matrix which reorders the columns of $\mathbf{X}$. Let $\mathbf{Z}$ be feasible to $\mathbf{X} = \mathbf{X}\mathbf{Z}$. Then $\tilde{\mathbf{Z}} = \mathbf{P}^\top \mathbf{Z} \mathbf{P}$ is feasible to $\tilde{\mathbf{X}} = \tilde{\mathbf{X}}\tilde{\mathbf{Z}}$. The EBD condition (1) guarantees that $f(\mathbf{Z}, \mathbf{X}) = f(\hat{\mathbf{Z}}, \hat{\mathbf{X}})$. Thus, $\hat{\mathbf{Z}}$ is equivalent to $\mathbf{Z}$ up to any reordering of the input data matrix $\mathbf{X}$. This is necessary for data clustering.

2. The EBD condition (2) is the key which enforces the solutions to (5) to be block diagonal under certain subspace assumption. From Theorem 1, we have $\mathbf{X} = \mathbf{XZ} = \mathbf{XZ}^B$. So the EBD condition (2) guarantees that $\mathbf{Z} = \mathbf{Z}^B$ when minimizing the objective. This will be more clear from the proof of Theorem 3.

3. The EBD condition (3) is actually not necessary to enforce the solutions to (5) to be block diagonal. But through the lens of this condition, we will see the connection between the structure of each block of the block diagonal solutions and the used objective $f$. Also, we find that many objectives in existing methods satisfy this condition.

The EBD conditions are not restrictive. Before giving the examples, we provide some useful properties discussing different types of functions that satisfy the EBD conditions.

**Proposition 1.** *If $f$ satisfies the EBD conditions (1)-(3) on $(\Omega, \Delta)$, then it does on $(\Omega_1, \Delta)$, where $\Omega_1 \subset \Omega$ and $\Omega_1 \neq \emptyset$.*

**Proposition 2.** *Assume that $f(\mathbf{Z}, \mathbf{X}) = \sum_{ij} g_{ij}(z_{ij})$, where $g_{ij}$ is a function defined on $\Omega_{ij}$, and it satisfies that $g_{ij}(z_{ij}) \geq 0$, $g_{ij}(z_{ij}) = 0$ if and only if $z_{ij} = 0$. Then $f$ satisfies the EBD conditions (1)-(3) on $(\Omega, \mathbb{R}^{D \times n})$, where $\Omega = \{\mathbf{Z} | z_{ij} \in \Omega_{ij}\}$.*

**Proposition 3.** *Assume that $f(\mathbf{Z}, \mathbf{X}) = \sum_j g_j([\mathbf{Z}]_{:,j}, \mathbf{X})$, where $g_j$ is a function defined on $(\Omega_j, \Delta)$. Assume that $\mathbf{X} = [\mathbf{X}_1, \mathbf{X}_2]$, $\mathbf{w} = [\mathbf{w}_1; \mathbf{w}_2] \in \Omega_j$, $\mathbf{w}^B = [\mathbf{w}_1; 0] \in \Omega_j$, and their dimensions are compatible. If $g_j$ satisfies the following conditions:*

*(1) $g_j(\mathbf{w}, \mathbf{X}) = g_j(\mathbf{P}^\top \mathbf{w}, \mathbf{XP})$, for any permutation matrix $\mathbf{P}$, $\mathbf{P}^\top \mathbf{w} \in \Omega_j$,*

*(2) $g_j(\mathbf{w}, \mathbf{X}) \geq g_j(\mathbf{w}^B, \mathbf{X})$, where the equality holds if and only if $\mathbf{w} = \mathbf{w}^B$,*

*(3) $g_j(\mathbf{w}^B, \mathbf{X}) = g_j(\mathbf{w}_1, \mathbf{X}_1)$,*

*then $f$ satisfies the EBD conditions (1)-(3) on $(\Omega, \Delta)$, where $\Omega = \{\mathbf{Z} | [\mathbf{Z}]_{:,j} \in \Omega_j\}$.*

**Proposition 4.** *Assume that $f(\mathbf{Z}, \mathbf{X}) = \sum_i g_i([\mathbf{Z}]_{i,:}, \mathbf{X})$, where $g_i$ is a function defined on $(\Omega_i, \Delta)$. Assume that $\mathbf{X} = [\mathbf{X}_1, \mathbf{X}_2]$, $\mathbf{w}^\top = [\mathbf{w}_1; \mathbf{w}_2]^\top \in \Omega_i$, $(\mathbf{w}^B)^\top = [\mathbf{w}_1; 0]^\top \in \Omega_i$, and their dimensions are compatible. If $g_i$ satisfies the following conditions:*

*(1) $g_i(\mathbf{w}^\top, \mathbf{X}) = g_i(\mathbf{w}^\top \mathbf{P}, \mathbf{XP})$, for any permutation matrix $\mathbf{P}$, $\mathbf{w}^\top \mathbf{P} \in \Omega_i$,*

*(2) $g_i(\mathbf{w}^\top, \mathbf{X}) \geq g_i((\mathbf{w}^B)^\top, \mathbf{X})$, where the equality holds if and only if $\mathbf{w} = \mathbf{w}^B$,*

*(3) $g_i((\mathbf{w}^B)^\top, \mathbf{X}) = g_i(\mathbf{w}_1^\top, \mathbf{X}_1)$,*

*then $f$ satisfies the EBD conditions (1)-(3) on $(\Omega, \Delta)$, where $\Omega = \{\mathbf{Z} | [\mathbf{Z}]_{i,:} \in \Omega_i\}$.*

**Proposition 5.** *If $f_i$ satisfies the EBD conditions (1)-(3) on $(\Omega_i, \Delta)$, $i = 1, \cdots, m$, then $\sum_i^m \lambda_i f_i$ (for positive $\lambda_i$) also satisfies the EBD conditions (1)-(3) on $(\Omega, \Delta)$ when $\Omega = \cap_{i=1}^m \Omega_i$ and $\Omega \neq \emptyset$.*

**Proposition 6.** *Assume that $f_1$ satisfies the EBD conditions (1)-(3) on $(\Omega_1, \Delta)$, $f_2$ satisfies the EBD conditions (1)-(3) on $(\Omega_2, \Delta)$ and $f_2(\mathbf{Z}, \mathbf{X}) \geq f_2(\mathbf{Z}^B, \mathbf{X})$, where $\mathbf{Z}, \mathbf{Z}^B$ and $\mathbf{X}$ are the same as those in Definition 3. Then, $f_1 + f_2$ satisfies the EBD conditions (1)-(3) on $(\Omega, \Delta)$ when $\Omega = \Omega_1 \cap \Omega_2$ and $\Omega \neq \emptyset$.*

**Theorem 2.** *Some functions of interest which satisfy the EBD conditions (1)-(3) are:*

| Function | $f(\mathbf{Z}, \mathbf{X})$ | $(\Omega, \Delta)$ |
|---|---|---|
| $\ell_0$- and $\ell_1$-norm | $\|\mathbf{Z}\|_0$ and $\|\mathbf{Z}\|_1$ | - |
| square of Frobenius norm | $\|\mathbf{Z}\|^2$ | - |
| elastic net | $\|\mathbf{Z}\|_1 + \lambda \|\mathbf{Z}\|^2$ | - |
| $\ell_{2,1}$-norm | $\|\mathbf{Z}\|_{2,1}$ | - |
| $\ell_{1,2}$-norm | $\|\mathbf{Z}\|_{1,2}$ | - |
| - | $\|\mathbf{Z}^\top \mathbf{Z}\|_1$ | $\Omega = \{\mathbf{Z} | \mathbf{Z} \geq 0\}$ |
| $\ell_1$+nuclear norm | $\|\mathbf{Z}\|_1 + \lambda \|\mathbf{Z}\|_*$ | - |
| trace Lasso | $\sum_j \|\mathbf{X}\text{Diag}([\mathbf{Z}]_{:,j})\|_*$ | $\Delta = \{\mathbf{X} | \forall j,$ $[\mathbf{X}]_{:,j} \neq 0\}$ |
| others | $\sum_{ij} \lambda_{ij} |z_{ij}|^{p_{ij}}$ | |

[1] $\Omega$ (resp. $\Delta$) is not specified if there has no restriction on $\mathbf{Z}$ (resp. $\mathbf{X}$).
[2] For the parameters, $\lambda > 0$, $\lambda_{ij} > 0$, $p_{ij} \geq 0$.

Theorem 2 gives some functions of interest which satisfy the EBD conditions. They can be verified by using Propositions 2-6. An intuitive verification is discussed as follows and the detailed proofs can be found in the supplementary material.

1. Proposition 2 verifies the EBD conditions of functions which are separable w.r.t. each element of a matrix, e.g., $\|\mathbf{Z}\|_0$, $\|\mathbf{Z}\|_1$, $\|\mathbf{Z}\|^2$ and $\sum_{ij} \lambda_{ij} |z_{ij}|^{p_{ij}}$.

2. Proposition 3 verifies the EBD conditions of functions which are separable w.r.t. each column of a matrix, e.g., $\|\mathbf{Z}\|_{2,1}$ and $\sum_i \|\mathbf{X}\text{Diag}([\mathbf{Z}]_{:,i})\|_*$.

3. Proposition 4 verifies the EBD conditions of functions which are separable w.r.t. each row of a matrix, e.g., $\|\mathbf{Z}\|_{1,2}$.

4. Proposition 5 shows that the function which is a positive linear combination of functions that satisfy the EBD conditions still satisfies the EBD conditions, e.g., $\|\mathbf{Z}\|_1 + \lambda \|\mathbf{Z}\|^2$ or more generally $\|\mathbf{Z}\|_0 + \lambda_1 \|\mathbf{Z}\|_1 + \lambda_2 \|\mathbf{Z}\|^2 + \lambda_3 \|\mathbf{Z}\|_{2,1} + \lambda_4 \|\mathbf{Z}\|_{1,2} + \lambda_5 \|\mathbf{Z}^\top \mathbf{Z}\|_1 + \lambda_6 \|\mathbf{Z}\|_* + \lambda_7 \sum_i \|\mathbf{X}\text{Diag}([\mathbf{Z}]_{:,i})\|_*$, where $\lambda_i > 0$. So Proposition 5 enlarges the family of such type of functions and shows that the EBD conditions are not restrictive.

5. Proposition 6 shows that $f_1 + f_2$ satisfies the EBD conditions (1)-(3) when $f_1$ satisfies the EBD conditions (1)-(3) and $f_2$ satisfies the EBD conditions (1)(3) and the first part of EBD condition (2). An example is $\|\mathbf{Z}\|_1 + \lambda \|\mathbf{Z}\|_*$. See more discussions about $\|\mathbf{Z}\|_*$ below.

There are also some interesting norms which do not satisfy the EBD conditions. For example, considering the infinity norm $\|\mathbf{Z}\|_\infty$, the EBD condition (1) holds while the other two do not. The nuclear norm $\|\mathbf{Z}\|_*$ satisfies the EBD condition (1)(3). But for the EBD condition (2), we only have (see Lemma 7.4 in [21])

$$\left\| \begin{bmatrix} \mathbf{Z}_1 & \mathbf{Z}_3 \\ \mathbf{Z}_4 & \mathbf{Z}_2 \end{bmatrix} \right\|_* \geq \left\| \begin{bmatrix} \mathbf{Z}_1 & 0 \\ 0 & \mathbf{Z}_2 \end{bmatrix} \right\|_* = \|\mathbf{Z}_1\|_* + \|\mathbf{Z}_2\|_*.$$

But the equality may hold when $\mathbf{Z}_3 \neq 0$ and $\mathbf{Z}_4 \neq 0$. A counterexample is that, when both $\mathbf{Z}$ and $\mathbf{Z}^B$ are positive semidefinite, $\|\mathbf{Z}\|_* = \sum_i \lambda_i(\mathbf{Z}) = \text{Tr}(\mathbf{Z}) = \text{Tr}(\mathbf{Z}^B) = \sum_i \lambda_i(\mathbf{Z}^B) = \|\mathbf{Z}^B\|_*$, where $\lambda_i(\mathbf{Z})$'s denote the eigenvalues of $\mathbf{Z}$. As will be seen in the proof of Theorem 3, this issue

makes the proof of the block diagonal property of LRR which uses the nuclear norm different from others. We instead use the uniqueness of the LRR solution to (4) to fix this issue.

**Theorem 3.** *Consider a collection of data points drawn from $k$ independent subspaces $\{\mathcal{S}_i\}_{i=1}^k$ of dimensions $\{d_i\}_{i=1}^k$. Let $\mathbf{X}_i \in \mathbb{R}^{D \times n_i}$ denote the data points in $\mathcal{S}_i$, $rank(\mathbf{X}_i) = d_i$ and $\sum_{i=1}^k n_i = n$. Let $\mathbf{X} = [\mathbf{X}_1, \cdots, \mathbf{X}_k] \in \Delta$, where $\Delta$ is a set consisting of matrices with nonzero columns. Considering problem (5), assume that $\{\mathbf{Z}|\mathbf{X} = \mathbf{XZ}\} \cap \Omega$ is nonempty and let $\mathbf{Z}^*$ be any optimal solution. If one of the following cases holds, Case I: $f$ satisfies the EBD condition (1)-(2) on $(\Omega, \Delta)$, Case II: $f$ satisfies the EBD condition (1) on $(\Omega, \Delta)$ and $\mathbf{Z}^*$ is the unique solution, then $\mathbf{Z}^*$ satisfies the block diagonal property, i.e.,*

$$\mathbf{Z}^* = \begin{bmatrix} \mathbf{Z}_1^* & 0 & \cdots & 0 \\ 0 & \mathbf{Z}_2^* & \cdots & 0 \\ \vdots & \vdots & \ddots & \vdots \\ 0 & 0 & \cdots & \mathbf{Z}_k^* \end{bmatrix}, \quad (8)$$

*with $\mathbf{Z}_i^* \in \mathbb{R}^{n_i \times n_i}$ corresponding to $\mathbf{X}_i$. Furthermore, if $f$ satisfies the EBD condition (1)-(3), then each block $\mathbf{Z}_i^*$ in (8) is optimal to the following problem:*

$$\min_{\mathbf{W}} f(\mathbf{W}, \mathbf{X}_i) \ \ s.t. \ \mathbf{X}_i = \mathbf{X}_i \mathbf{W}, \mathbf{W} \in \Omega. \quad (9)$$

*Proof.* First, by the EBD condition (1), $f(\mathbf{Z}, \mathbf{X}) = f(\mathbf{P}^\top \mathbf{ZP}, \mathbf{XP})$ holds for any permutation $\mathbf{P}$. This guarantees that the learned $\mathbf{Z}^*$ based on $\mathbf{X}$ by solving (5) is equivalent to $\mathbf{P}^\top \mathbf{Z}^* \mathbf{P}$ based on $\mathbf{XP}$. So we only need to discuss the structure of $\mathbf{Z}^*$ based on the ordered input data matrix $\mathbf{X} = [\mathbf{X}_1, \cdots, \mathbf{X}_k]$.

For any optimal solution $\mathbf{Z}^* \in \Omega$ to problem (5), we decompose it into two parts $\mathbf{Z}^* = \mathbf{Z}^B + \mathbf{Z}^C$, where $\mathbf{Z}^B$ and $\mathbf{Z}^C$ are of the forms in (7). Then, by Theorem 1, we have $\mathbf{XZ}^B = \mathbf{X}$ and $\mathbf{XZ}^C = 0$. This combines the EBD conditions, which implies that $\mathbf{Z}^B$ is feasible to (5). By the EBD conditions (2), we have $f(\mathbf{Z}^*, \mathbf{X}) \geq f(\mathbf{Z}^B, \mathbf{X})$. On the other hand, $\mathbf{Z}^*$ is optimal to (5), thus we have $f(\mathbf{Z}^*, \mathbf{X}) \leq f(\mathbf{Z}^B, \mathbf{X})$. Therefore, $f(\mathbf{Z}^*, \mathbf{X}) = f(\mathbf{Z}^B, \mathbf{X})$. In Case I, by the EBD condition (2), we have $\mathbf{Z}^* = \mathbf{Z}^B$. The same result holds in Case II. Hence, $\mathbf{Z}^* = \mathbf{Z}^B$ satisfies the block diagonal property in both cases.

If the EBD condition (3) is further satisfied, we have $f(\mathbf{Z}^*, \mathbf{X}) = \sum_{i=1}^k f(\mathbf{Z}_i^*, \mathbf{X}_i)$, which is separable. By the block diagonal structure of $\mathbf{Z}^*$, $\mathbf{X} = \mathbf{XZ}^*$ is equivalent to $\mathbf{X}_i = \mathbf{X}_i \mathbf{Z}_i^*$, $i = 1, \cdots, k$. Hence, both the objectives and constraints of (5) are separable and thus problem (5) is equivalent to problem (9) for all $i = 1, \cdots, k$. This guarantees the same solutions of (5) and (9). ☐

We have the following remarks for Theorem 3:

1. Theorem 3 gives a general guarantee of the block diagonal property for the solutions to (5) based on the EBD conditions. By Theorem 2, the block diagonal properties of existing methods (except LRR) in Table 1 are special cases of Theorem 3 (Case I). Note that some existing models, e.g., SSC, have a constraint $\text{diag}(\mathbf{Z}) = 0$. This does not affect the EBD conditions due to Proposition 1. Actually, additional proper constraints can be introduced in (5) if necessary and the block diagonal property still holds.

2. The nuclear norm used in LRR does not satisfy the EBD condition (2). Fortunately, the LRR model (4) has a unique solution [21]. Thus the block diagonal property of LRR is another special case of Theorem 3 (Case II). If we choose $\Omega = \{\mathbf{Z}|\mathbf{X} = \mathbf{XZ}\}$, then the nuclear norm satisfies the EBD conditions (1)(2) on $(\Omega, \mathbb{R}^{d \times n})$ due to the uniqueness of LRR. So, in some cases, the Case II can be regarded as a special case of Case I in Theorem 3.

3. The SSQP method [40] achieves the solution obeying the block diagonal property under the orthogonal subspace assumption. However, the EBD conditions and Theorem 3 show that the weaker independent subspace assumption is enough. Actually, if the subspaces are orthogonal, $\mathbf{X}^\top \mathbf{X}$ already obeys the block diagonal property.

4. Theorem 3 not only provides the block diagonal property guarantee of $\mathbf{Z}^*$ (there are no connections between-subspaces), but also shows what property each block has (the property of the connections within-subspace). Let us take the SSC model as an example. The $i$-th block $\mathbf{Z}_i^*$ of $\mathbf{Z}^*$, which is optimal to (3), is the minimizer to

$$\mathbf{Z}_i^* = \arg\min_{\mathbf{W}} \|\mathbf{W}\|_1 \ \ \text{s.t.} \ \mathbf{X}_i = \mathbf{X}_i \mathbf{W}, \text{diag}(\mathbf{W}) = 0.$$

So SSC not only finds a sparse representation between-subspaces but also within-subspace. Hence, each $\mathbf{Z}_i^*$ may be too sparse (not fully connected) especially when the columns of $\mathbf{X}_i$ are highly correlated. This perspective provides an intuitive interpretation of the graph connectivity issue in SSC.

5. Theorem 3 not only provides a good summary of existing methods, but also provides the general motivation for designing new subspace clustering methods as the EBD conditions are easy to verify by using Proposition 1-6.

## 3 SUBSPACE CLUSTERING BY BLOCK DIAGONAL REPRESENTATION

Theorem 3 shows that it is not difficult to find a solution obeying the block diagonal property under the independent subspaces assumption as the EBD conditions are not restrictive. Usually, the solution is far from being $k$-block diagonal since the independent subspaces assumption does not hold due to data noise. The more direct method [11] enforces the representation coefficient matrix to be $k$-block diagonal with exact $k$ connected blocks. However, in practice, the $k$-block diagonal affinity matrix is not necessary for correct clustering when using spectral clustering. Similar phenomenons are observed in the pursuits of sparsity and low-rankness. The sparsity (or low-rankness) is widely used as a prior in many applications, but the exact sparsity (or rank) is not (necessarily) known. So the $\ell_1$-norm (or nuclear norm) is very widely used as a regularizer to encourage the solution to be sparse (or low-rank). Now, considering the $k$-block diagonal matrix, which is another interesting structure, what is the corresponding regularizer?

In this section, we will propose a simple *block diagonal regularizer* for pursuing such an interesting structure. By using this regularizer, we then propose a direct subspace clustering subspace method, termed *Block Diagonal Representation (BDR)*. We will also propose an efficient solver and provide the convergence guarantee.

### 3.1 Block Diagonal Regularizer

In this work, we say that a matrix is $k$-block diagonal if it has at least $k$ connected components (blocks). Such a concept is somewhat ambiguous. For example, consider the following matrix

$$\mathbf{B} = \begin{bmatrix} \mathbf{B}_0 & 0 & 0 \\ 0 & \mathbf{B}_0 & 0 \\ 0 & 0 & \mathbf{B}_0 \end{bmatrix}, \text{ where } \mathbf{B}_0 = \begin{bmatrix} 1 & 0 \\ -1 & 1 \end{bmatrix} \quad (10)$$

is fully connected. We can say that $\mathbf{B}$ is 3-block diagonal (this is what we expect intuitively). But by the definition, we can also say that it is 1- or 2-block diagonal. Thus, we need a more precise way to characterize the number of connected components.

Assume that $\mathbf{B}$ is an affinity matrix, i.e., $\mathbf{B} \geq 0$ and $\mathbf{B} = \mathbf{B}^\top$, the corresponding Laplacian matrix, denoted as $\mathbf{L_B}$, is defined as

$$\mathbf{L_B} = \text{Diag}(\mathbf{B1}) - \mathbf{B}.$$

The number of connected components of $\mathbf{B}$ is related to the spectral property of the Laplacian matrix.

**Theorem 4.** *[39, Proposition 4] For any $\mathbf{B} \geq 0$, $\mathbf{B} = \mathbf{B}^\top$, the multiplicity $k$ of the eigenvalue 0 of the corresponding Laplacian matrix $\mathbf{L_B}$ equals the number of connected components (blocks) in $\mathbf{B}$.*

For any affinity matrix $\mathbf{B} \in \mathbb{R}^{n \times n}$, let $\lambda_i(\mathbf{L_B})$, $i = 1, \cdots, n$, be the eigenvalues of $\mathbf{L_B}$ in the decreasing order. It is known that $\mathbf{L_B} \succeq 0$ and thus $\lambda_i(\mathbf{L_B}) \geq 0$ for all $i$. Then, by Theorem 4, $\mathbf{B}$ has $k$ connected components if and only if

$$\lambda_i(\mathbf{L_B}) \begin{cases} > 0, & i = 1, \cdots, n-k, \\ = 0, & i = n-k+1, \cdots, n. \end{cases} \quad (11)$$

Motivated by such a property, we define the $k$-block diagonal regularizer as follows.

**Definition 4.** *($k$-block diagonal regularizer) For any affinity matrix $\mathbf{B} \in \mathbb{R}^{n \times n}$, the $k$-block diagonal regularizer is defined as the sum of the $k$ smallest eigenvalues of $\mathbf{L_B}$, i.e.,*

$$\|\mathbf{B}\|_{\boxed{k}} = \sum_{i=n-k+1}^{n} \lambda_i(\mathbf{L_B}). \quad (12)$$

It can be seen that $\|\mathbf{B}\|_{\boxed{k}} = 0$ is equivalent to the fact that the affinity matrix $\mathbf{B}$ is $k$-block diagonal. So $\|\mathbf{B}\|_{\boxed{k}}$ can be regarded as the block diagonal matrix structure induced regularizer.

It is worth mentioning that (11) is equivalent to $\text{rank}(\mathbf{L_B}) = n - k$. One may consider using $\text{rank}(\mathbf{L_B})$ as the $k$-block diagonal regularizer. However, this is not a good choice. The reason is that the number of data points $n$ is usually much larger than the number of clusters $k$ and thus $\mathbf{L_B}$ is of high rank. It is generally unreasonable to find a high rank matrix by minimizing $\text{rank}(\mathbf{L_B})$. More importantly, it is not able to control the targeted number of blocks, which is important in subspace clustering. Another choice is the convex relaxation $\|\mathbf{L_B}\|_*$, but it suffers from the same issues.

It is interesting that the sparse minimization in the SSC model (3) is equivalent to minimizing $\|\mathbf{L_B}\|_*$. Indeed,

$$\|\mathbf{L_B}\|_* = \text{Tr}(\mathbf{L_B}) = \text{Tr}(\text{Diag}(\mathbf{B1}) - \mathbf{B})$$
$$= \|\mathbf{B}\|_1 - \|\text{diag}(\mathbf{B})\|_1,$$

where we use the facts that $\mathbf{B} = \mathbf{B}^\top$, $\mathbf{B} \geq 0$ and $\mathbf{L_B} \succeq 0$. So, the SSC model (3) is equivalent to

$$\min_{\mathbf{Z}} \|\mathbf{L_B}\|_*$$
$$\text{s.t. } \mathbf{X} = \mathbf{XZ}, \text{diag}(\mathbf{Z}) = 0, \ \mathbf{B} = (|\mathbf{Z}| + |\mathbf{Z}^\top|)/2.$$

This perspective shows that the approximation of the block diagonal matrix by using sparse prior in SSC is loose. In contrast, our proposed $k$-block diagonal regularizer (12) not only directly encourages the matrix to be block diagonal, but is also able to control the number of blocks, which is important for subspace clustering. A disadvantage is that the $k$-block diagonal regularizer is nonconvex.

### 3.2 Block Diagonal Representation

With the proposed $k$-block diagonal regularizer at hand, we now propose the Block Diagonal Representation (BDR) method for subspace clustering. We directly consider the BDR model for handling data with noise

$$\min_{\mathbf{B}} \frac{1}{2}\|\mathbf{X} - \mathbf{XB}\|^2 + \gamma \|\mathbf{B}\|_{\boxed{k}},$$
$$\text{s.t. } \text{diag}(\mathbf{B}) = 0, \mathbf{B} \geq 0, \mathbf{B} = \mathbf{B}^\top,$$

where $\gamma > 0$ and we simply require the representation matrix $\mathbf{B}$ to be nonnegative and symmetric, which are necessary properties for defining the block diagonal regularizer on $\mathbf{B}$. But the restrictions on $\mathbf{B}$ will limit its representation capability. We alleviate this issue by introducing an intermediate term

$$\min_{\mathbf{Z},\mathbf{B}} \frac{1}{2}\|\mathbf{X} - \mathbf{XZ}\|^2 + \frac{\lambda}{2}\|\mathbf{Z} - \mathbf{B}\|^2 + \gamma \|\mathbf{B}\|_{\boxed{k}},$$
$$\text{s.t. } \text{diag}(\mathbf{B}) = 0, \mathbf{B} \geq 0, \mathbf{B} = \mathbf{B}^\top. \quad (13)$$

The above two models are equivalent when $\lambda > 0$ is sufficiently large. As will be seen in Section 3.3, another benefit of the term $\frac{\lambda}{2}\|\mathbf{Z} - \mathbf{B}\|^2$ is that it makes the subproblems involved in updating $\mathbf{Z}$ and $\mathbf{B}$ strongly convex and thus the solutions are unique and stable. This also makes the convergence analysis easy.

**Example 1.** We give an intuitive example to illustrate the effectiveness of BDR. We generate a data matrix $\mathbf{X} = [\mathbf{X}_1, \mathbf{X}_2, \cdots, \mathbf{X}_k]$ with its columns sampled from $k$ subspaces without noise. We generate $k = 5$ subspaces $\{\mathcal{S}_i\}_{i=1}^k$ whose bases $\{\mathbf{U}_i\}_{i=1}^k$ are computed by $\mathbf{U}_{i+1} = \mathbf{T}\mathbf{U}_i$, $1 \leq i \leq k$, where $\mathbf{T}$ is a random rotation matrix and $\mathbf{U}_1 \in \mathbb{R}^{D \times r}$ is a random orthogonal matrix. We set $D = 30$ and $r = 5$. For each subpace, we sample $n_i = 50$ data vectors by $\mathbf{X}_i = \mathbf{U}_i\mathbf{Q}_i$, $1 \leq i \leq k$, with $\mathbf{Q}_i$ being an $r \times n_i$ i.i.d. $\mathcal{N}(0,1)$ matrix. So we have $\mathbf{X} \in \mathbb{R}^{D \times n}$, where $n = 250$. Each column of $\mathbf{X}$ is normalized to have a unit length. We then solve (13) to achieve $\mathbf{Z}$ and $\mathbf{B}$ (we set $\lambda = 10$ and $\gamma = 5$). Note that the generated data matrix $\mathbf{X}$ is noise free. So we also compute the shape interaction matrix $\mathbf{V}\mathbf{V}^\top$ (here $\mathbf{V}$ is from the skinny SVD of $\mathbf{X} = \mathbf{U}\mathbf{S}\mathbf{V}^\top$), which is the solution to the LRR model (4), for comparison. We plot

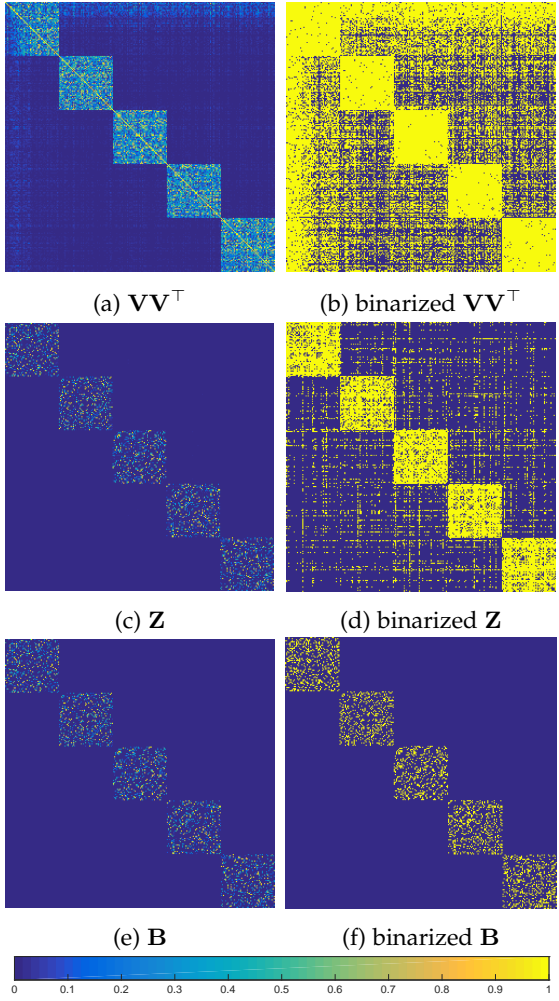

(a) $\mathbf{V}\mathbf{V}^\top$      (b) binarized $\mathbf{V}\mathbf{V}^\top$

(c) $\mathbf{Z}$      (d) binarized $\mathbf{Z}$

(e) $\mathbf{B}$      (f) binarized $\mathbf{B}$

Fig. 2: Plots of the shape interaction matrix $\mathbf{V}\mathbf{V}^\top$, $\mathbf{Z}$ and $\mathbf{B}$ from BDR and their binarized versions respectively for Example 1.

$\mathbf{V}\mathbf{V}^\top$, $\mathbf{Z}$ and $\mathbf{B}$ and their binarized versions in Figure 2. The binarization $\hat{\mathbf{Z}}$ of a matrix $\mathbf{Z}$ is defined as

$$\hat{\mathbf{Z}}_{ij} = \begin{cases} 0, & \text{if } |\mathbf{Z}_{ij}| <= \tau, \\ 1, & \text{otherwise}, \end{cases}$$

where we use $\tau = 10^{-3}$. From Figure 2, it can be seen that both $\mathbf{V}\mathbf{V}^\top$ and its binarized version are very dense and neither of them obeys the block diagonal property. This implies that the generated subspaces are not independent, though the sampled data points are noise free. In contrast, the obtained $\mathbf{B}$ by our BDR and its binarized version are not only $k$-block diagonal but they also obey the block diagonal property (this observation does not depend on the choice of the binarization parameter $\tau$). This experiment clearly shows the effectiveness of the proposed $k$-block diagonal regularizer for pursuing a solution obeying the block diagonal property in the case that the independent subspaces assumption is violated. Moreover, we observe that $\mathbf{Z}$ is close to but denser than $\mathbf{B}$. From the binarized version, we see that $\mathbf{Z}$ is not a $k$-block diagonal matrix. However, when applying the spectral clustering algorithm on $\mathbf{Z}$ and $\mathbf{B}$, we find that both lead to correct clustering while $\mathbf{V}\mathbf{V}^\top$ does not. This shows the robustness of spectral clustering to the

---

**Algorithm 1** Solve (14) by Alternating Minimization

**Initialize:** $k = 0$, $\mathbf{W}^k = 0$, $\mathbf{Z}^k = 0$, $\mathbf{B}^k = 0$.

**while** not converged **do**
1) Compute $\mathbf{W}^{k+1}$ by solving (15);
2) Compute $\mathbf{Z}^{k+1}$ by solving (16);
3) Compute $\mathbf{B}^{k+1}$ by solving (17);
4) $k = k + 1$.

**end while**

---

affinity matrix which is not but "close to" $k$-block diagonal. When $\gamma$ is relatively smaller, we observe that $\mathbf{B}$ may not be $k$-block diagonal, but it still leads to correct clustering. This shows that, for the subspace clustering problem, the soft block diagonal regularizer is more flexible than the hard constraint in [11].

### 3.3 Optimization of BDR

We show how to solve the nonconvex problem (13). The key challenge lies in the nonconvex term $\|\mathbf{B}\|_{\boxed{k}}$. We introduce an interesting property about the sum of eigenvalues by Ky Fan to reformulate $\|\mathbf{B}\|_{\boxed{k}}$.

**Theorem 5.** *[9, p. 515] Let $\mathbf{L} \in \mathbb{R}^{n \times n}$ and $\mathbf{L} \succeq 0$. Then*

$$\sum_{i=n-k+1}^{n} \lambda_i(\mathbf{L}) = \min_{\mathbf{W}} \langle \mathbf{L}, \mathbf{W} \rangle, \ \ s.t. \ 0 \preceq \mathbf{W} \preceq \mathbf{I}, \ Tr(\mathbf{W}) = k.$$

Then, we can reformulate $\|\mathbf{B}\|_{\boxed{k}}$ as a convex program

$$\|\mathbf{B}\|_{\boxed{k}} = \min_{\mathbf{W}} \langle \mathbf{L}_{\mathbf{B}}, \mathbf{W} \rangle, \ \ s.t. \ 0 \preceq \mathbf{W} \preceq \mathbf{I}, \ Tr(\mathbf{W}) = k.$$

So (13) is equivalent to

$$\min_{\mathbf{Z}, \mathbf{B}, \mathbf{W}} \frac{1}{2}\|\mathbf{X} - \mathbf{X}\mathbf{Z}\|^2 + \frac{\lambda}{2}\|\mathbf{Z} - \mathbf{B}\|^2 + \gamma\langle \text{Diag}(\mathbf{B}\mathbf{1}) - \mathbf{B}, \mathbf{W} \rangle$$
$$\text{s.t. } \text{diag}(\mathbf{B}) = 0, \mathbf{B} \geq 0, \mathbf{B} = \mathbf{B}^\top, \quad (14)$$
$$0 \preceq \mathbf{W} \preceq \mathbf{I}, \text{Tr}(\mathbf{W}) = k.$$

There are 3 blocks of variables in problem (14). We observe that $\mathbf{W}$ is independent from $\mathbf{Z}$, thus we can group them as a super block $\{\mathbf{W}, \mathbf{Z}\}$ and treat $\{\mathbf{B}\}$ as the other block. Then (14) can be solved by alternating updating $\{\mathbf{W}, \mathbf{Z}\}$ and $\{\mathbf{B}\}$.

First, fix $\mathbf{B} = \mathbf{B}^k$, and update $\{\mathbf{W}^{k+1}, \mathbf{Z}^{k+1}\}$ by

$$\{\mathbf{W}^{k+1}, \mathbf{Z}^{k+1}\} = \arg\min_{\mathbf{W}, \mathbf{Z}} \frac{1}{2}\|\mathbf{X} - \mathbf{X}\mathbf{Z}\|^2 + \frac{\lambda}{2}\|\mathbf{Z} - \mathbf{B}\|^2$$
$$+ \gamma\langle \text{Diag}(\mathbf{B}\mathbf{1}) - \mathbf{B}, \mathbf{W} \rangle$$
$$\text{s.t. } 0 \preceq \mathbf{W} \preceq \mathbf{I}, \text{Tr}(\mathbf{W}) = k.$$

This is equivalent to updating $\mathbf{W}^{k+1}$ and $\mathbf{Z}^{k+1}$ separably by

$$\mathbf{W}^{k+1} = \arg\min_{\mathbf{W}} \langle \text{Diag}(\mathbf{B}\mathbf{1}) - \mathbf{B}, \mathbf{W} \rangle,$$
$$\text{s.t. } 0 \preceq \mathbf{W} \preceq \mathbf{I}, \text{Tr}(\mathbf{W}) = k, \quad (15)$$

and

$$\mathbf{Z}^{k+1} = \arg\min_{\mathbf{Z}} \frac{1}{2}\|\mathbf{X} - \mathbf{X}\mathbf{Z}\|^2 + \frac{\lambda}{2}\|\mathbf{Z} - \mathbf{B}\|^2. \quad (16)$$

TABLE 2: Clustering errors (%) of different algorithms on the Hopkins 155 database with the $2F$-dimensional data points.

| method | SCC | SSC | LRR | LSR | S³C | BDR-B | BDR-Z |
|--------|-----|-----|-----|-----|-----|-------|-------|
| 2 motions | | | | | | | |
| mean | 2.46 | 1.52 | 3.65 | 3.24 | 1.73 | 1.00 | **0.95** |
| median | 0.00 | 0.00 | 0.22 | 0.00 | 0.00 | 0.00 | 0.00 |
| 3 motions | | | | | | | |
| mean | 11.00 | 4.40 | 9.40 | 5.94 | 4.76 | 1.95 | **0.85** |
| median | 1.63 | 1.63 | 3.99 | 2.05 | 0.93 | 0.21 | 0.21 |
| All | | | | | | | |
| mean | 4.39 | 2.18 | 4.95 | 3.85 | 2.41 | 1.22 | **0.93** |
| median | 0.00 | 0.00 | 0.53 | 0.45 | 0.00 | 0.00 | 0.00 |

TABLE 3: Clustering errors (%) of different algorithms on the Hopkins 155 database with the $4k$-dimensional data points by applying PCA.

| method | SCC | SSC | LRR | LSR | S³C | BDR-B | BDR-Z |
|--------|-----|-----|-----|-----|-----|-------|-------|
| 2 motions | | | | | | | |
| mean | 3.58 | 1.83 | 4.22 | 3.35 | 1.81 | 1.26 | **1.04** |
| median | 0.00 | 0.00 | 0.29 | 0.29 | 0.00 | 0.00 | 0.00 |
| 3 motions | | | | | | | |
| mean | 7.11 | 4.40 | 9.43 | 6.13 | 5.01 | **1.22** | **1.22** |
| median | 0.47 | 0.56 | 3.70 | 2.05 | 1.06 | 0.21 | 0.20 |
| All | | | | | | | |
| mean | 4.37 | 2.41 | 5.40 | 3.97 | 2.53 | 1.25 | **1.08** |
| median | 00.00 | 0.00 | 0.53 | 0.53 | 0.00 | 0.00 | 0.00 |

Second, fix $\mathbf{W} = \mathbf{W}^{k+1}$ and $\mathbf{Z} = \mathbf{Z}^{k+1}$, and update $\mathbf{B}$ by

$$\mathbf{B}^{k+1} = \arg\min_{\mathbf{B}} \frac{\lambda}{2}\|\mathbf{Z} - \mathbf{B}\|^2 + \gamma\langle \text{Diag}(\mathbf{B1}) - \mathbf{B}, \mathbf{W}\rangle$$
$$\text{s.t. } \text{diag}(\mathbf{B}) = 0, \mathbf{B} \geq 0, \mathbf{B} = \mathbf{B}^\top. \quad (17)$$

Note that the above three subproblems are convex and have closed form solutions. For (15), $\mathbf{W}^{k+1} = \mathbf{U}\mathbf{U}^\top$, where $\mathbf{U} \in \mathbb{R}^{n \times k}$ consist of $k$ eigenvectors associated with the $k$ smallest eigenvalues of $\text{Diag}(\mathbf{B1}) - \mathbf{B}$. For (16), it is obvious that

$$\mathbf{Z}^{k+1} = (\mathbf{X}^\top\mathbf{X} + \lambda\mathbf{I})^{-1}(\mathbf{X}^\top\mathbf{X} + \lambda\mathbf{B}). \quad (18)$$

For (17), it is equivalent to

$$\mathbf{B}^{k+1} = \arg\min_{\mathbf{B}} \frac{1}{2}\left\|\mathbf{B} - \mathbf{Z} + \frac{\gamma}{\lambda}(\text{diag}(\mathbf{W})\mathbf{1}^\top - \mathbf{W})\right\|^2$$
$$\text{s.t. } \text{diag}(\mathbf{B}) = 0, \mathbf{B} \geq 0, \mathbf{B} = \mathbf{B}^\top. \quad (19)$$

This problem has a closed form solution given as follows.

**Proposition 7.** *Let* $\mathbf{A} \in \mathbb{R}^{n \times n}$. *Define* $\hat{\mathbf{A}} = \mathbf{A} - Diag(diag(\mathbf{A}))$. *Then the solution to the following problem*

$$\min_{\mathbf{B}} \frac{1}{2}\|\mathbf{B} - \mathbf{A}\|^2, \text{ s.t. } diag(\mathbf{B}) = 0, \mathbf{B} \geq 0, \mathbf{B} = \mathbf{B}^\top, \quad (20)$$

*is given by* $\mathbf{B}^* = \left[(\hat{\mathbf{A}} + \hat{\mathbf{A}}^\top)/2\right]_+$.

The whole procedure of the alternating minimization solver for (14) is given in Algorithm 1. We denote the objective of (14) as $f(\mathbf{Z}, \mathbf{B}, \mathbf{W})$. Let $S_1 = \{\mathbf{B}|\text{diag}(\mathbf{B}) = 0, \mathbf{B} \geq 0, \mathbf{B} = \mathbf{B}^\top\}$ and $S_2 = \{\mathbf{W}|0 \preceq \mathbf{W} \preceq \mathbf{I}, \text{Tr}(\mathbf{W}) = k\}$. Denote the indicator functions of $S_1$ and $S_2$ as $\iota_{S_1}(\mathbf{B})$ and $\iota_{S_2}(\mathbf{W})$, respectively. We give the convergence guarantee of Algorithm 1 for nonconvex BDR problem.

**Proposition 8.** *The sequence* $\{\mathbf{W}^k, \mathbf{Z}^k, \mathbf{B}^k\}$ *generated by Algorithm 1 has the following properties:*
*(1) The objective* $f(\mathbf{Z}^k, \mathbf{B}^k, \mathbf{W}^k) + \iota_{S_1}(\mathbf{B}^k) + \iota_{S_2}(\mathbf{W}^k)$ *is monotonically decreasing. Indeed,*

$$f(\mathbf{Z}^{k+1}, \mathbf{B}^{k+1}, \mathbf{W}^{k+1}) + \iota_{S_1}(\mathbf{B}^{k+1}) + \iota_{S_2}(\mathbf{W}^{k+1})$$
$$\leq f(\mathbf{Z}^k, \mathbf{B}^k, \mathbf{W}^k) + \iota_{S_1}(\mathbf{B}^k) + \iota_{S_2}(\mathbf{W}^k)$$
$$- \frac{\lambda}{2}\left\|\mathbf{Z}^{k+1} - \mathbf{Z}^k\right\|^2 - \frac{\lambda}{2}\left\|\mathbf{B}^{k+1} - \mathbf{B}^k\right\|^2;$$

*(2)* $\mathbf{Z}^{k+1} - \mathbf{Z}^k \to 0$, $\mathbf{B}^{k+1} - \mathbf{B}^k \to 0$ *and* $\mathbf{W}^{k+1} - \mathbf{W}^k \to 0$;
*(3) The sequences* $\{\mathbf{Z}^k\}$, $\{\mathbf{B}^k\}$ *and* $\{\mathbf{W}^k\}$ *are bounded.*

**Theorem 6.** *The sequence* $\{\mathbf{W}^k, \mathbf{Z}^k, \mathbf{B}^k\}$ *generated by Algorithm 1 has at least one limit point and any limit point* $(\mathbf{Z}^*, \mathbf{B}^*, \mathbf{W}^*)$ *of* $\{\mathbf{Z}^k, \mathbf{B}^k, \mathbf{W}^k\}$ *is a stationary point of (14).*

Please refer to the supplementary material for the proof of the above theorem. Generally, our proposed solver in Algorithm 1 for the nonconvex BDR model is simple. The convergence guarantee in Theorem 6 for Algorithm 1 is practical as there have no unverifiable assumptions.

### 3.4 Subspace Clustering Algorithm

We give the procedure of BDR for subspace clustering as previous works [10], [20], [26]. Given the data matrix $\mathbf{X}$, we obtain the representation matrix $\mathbf{Z}$ and $\mathbf{B}$ by solving the proposed BDR problem (13) by Algorithm 1. Both of them can be used to infer the data clustering. The affinity matrix can be defined as $\mathbf{W} = (|\mathbf{Z}| + |\mathbf{Z}^\top|)/2$ or $\mathbf{W} = (|\mathbf{B}| + |\mathbf{B}^\top|)/2$, and then the traditional spectral clustering [31] is applied on $\mathbf{W}$ to group the data points into $k$ groups. As will be seen in the experiments, the clustering performance on $\mathbf{Z}$ and $\mathbf{B}$ is comparable.

It is worth mentioning that our BDR requires to know the number of subspaces $k$ when computing the affinity matrix and using the spectral clustering to achieve the final result. Such a requirement is necessary for all the spectral-type subspace clustering methods, e.g., [10], [20], [26], though it is only used in the spectral clustering step. If the number of subspaces is not known, some other techniques can be used for the estimation, e.g., [4], [20]. This work only focuses on the case that the number of subspaces is known.

## 4 EXPERIMENTS

In this section, we conduct several experiments on real datasets to demonstrate the effectiveness of our BDR. The compared methods include SCC [7], SSC [10], LRR [20], LSR [26], S³C [19], BDR-B (our BDR model by using $\mathbf{B}$) and BDR-Z (our BDR model by using $\mathbf{Z}$). For the existing methods, we use the codes released by the authors. We test on three datasets: Hopkins 155 database [36] for motion segmentation, Extended Yale B [18] for face clustering and MNIST [13] for handwritten digit clustering. For all the compared methods, we tune the parameters (for some methods, we use the parameters which are given in their codes for some datasets) and use the ones which achieve the best results in most cases for each dataset. Note that BDR-B and BDR-Z use the same parameters[3].

---

3. We will release the codes of our BDR and the used datasets soon.

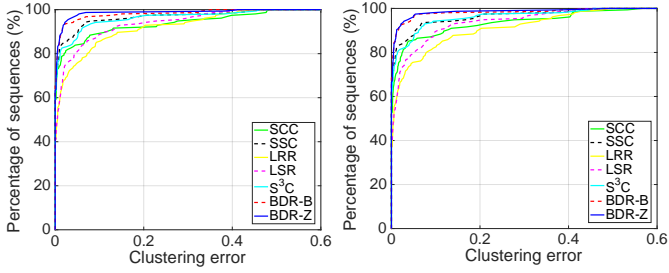

**TABLE 4:** The mean clustering errors (%) of 155 sequences on Hopkins 155 dataset by state-of-the-art methods.

| LSA [43] | SSC [10] | LRR [20] | LatLRR [22] | LSR [26] |
|----------|----------|----------|-------------|----------|
| 4.52 | 2.18 | 1.59 | **0.85** | 1.71 |
| CASS [23] | SMR [14] | BD-SSC [11] | BD-LRR [11] | BDR-Z |
| 1.47 | 1.13 | 1.68 | 0.97 | 0.93 |

Fig. 3: Percentage of sequences for which the clustering error is less than or equal to a given percentage of misclassification. Left: $2F$-dimensional data. Right: $4n$-dimensional data.

For the performance evaluation, we use the usual clustering error defined as follows

$$\text{clustering error} = 1 - \frac{1}{n}\sum_{i=1}^{n}\delta(p_i, \text{map}(q_i)), \quad (21)$$

where $p_i$ and $q_i$ represent the output label and the ground truth one of the $i$-th point respectively, $\delta(x, y) = 1$ if $x = y$, and $\delta(x, y) = 0$ otherwise, and $\text{map}(q_i)$ is the best mapping function that permutes clustering labels to match the ground truth labels.

## 4.1 Motion Segmentation

We consider the application of subspace clustering for motion segmentation. It refers to the problem of segmenting a video sequence with multiple rigidly moving objects into multiple spatiotemporal regions that correspond to the different motions in the scene. The coordinates of the points in trajectories of one moving object form a low dimensional subspace. Thus, the motion segmentation problem can be solved via performing subspace clustering on the trajectory spatial coordinates. We test on the widely used Hopkins 155 database [36]. It consists of 155 video sequences, where 120 of the videos have two motions and 35 of the videos have three motions. The feature trajectories of each video can be well modeled as data points that approximately lie in a union of linear subspaces of dimension at most 4 [10]. Each sequence is a sole dataset (i.e., data matrix $\mathbf{X}$) and so there are in total 155 subspace clustering tasks.

We consider two settings to construct the data matrix $\mathbf{X}$ for each sequence: (1) use the original $2F$-dimensional feature trajectories, where $F$ is the number of frames of the video sequence; (2) project the data matrix into $4k$-dimensional subspace, where $k$ is the number of subspaces, by using PCA. Most of the compared methods are spectral-type methods, except SCC. For spectral-type methods, they used different post-processing on the learned affinity matrices when using spectral clustering. We first consider the same setting as [10] which defines the affinity matrix by $\mathbf{W} = (|\mathbf{Z}| + |\mathbf{Z}^\top|)/2$, where $\mathbf{Z}$ is the learned representation coefficient matrix, and no additional complex post-processing is performed. In the Hopkins 155 database, there are 120 videos of two motions and 35 videos of three motions. So we report the mean and median of the clustering errors of these videos. Table 2 and Table 3 report the clustering errors of applying the compared methods on the dataset when we use the original $2F$-dimensional feature trajectories and when we project the data into a $4k$-dimensional subspace using PCA, respectively. Figure 3 gives the percentage of sequences for which the clustering error is less than or equal to a given percentage of misclassification. Furthermore, consider that many subspace clustering methods achieve state-of-the-art performance on the Hopkins 155 database by using different techniques for pre-processing and post-processing. So we give a direct performance comparison of the subspace clustering methods with their reported settings on all 155 sequences in Table 4. Based on these results, we have the following observations:

- From Table 2 and Table 3, it can be seen that our BDR-B and BDR-Z achieve close performance and both outperform the existing methods in both settings, though many existing methods already perform very well. Considering that the reported results are the means of the clustering errors of many sequences, the improvements (from the existing best result 2.18% to our 0.93% in Table 2 and from the existing best result 2.41% to our 1.08% in Table 3) by our BDR-B and BDR-Z are significant.

- From Figure 3, it can be seen that there are many more sequences which are almost correctly segmented by our BDR-B abd BDR-Z than existing methods. This demonstrates that the improvements over existing methods by our methods are achieved on most of the sequences.

- For most methods, the clustering performance using the $2F$-dimensional feature trajectories in Table 2 is slightly better than using the $4k$-dimensional PCA projections in Table 3. This implies that the feature trajectories of $k$ motions in a video almost perfectly lie in a $4k$-dimensional linear subspace of the $2F$-dimensional ambient space.

- From Table 4, it can be seen that our BDR-Z performed on the $2F$-dimensional data points still outperforms many existing state-of-the-art methods which use various post-processing techniques. LatLRR [22] is slightly better than our method. But it requires much more complex pre-processing and post-processing, and much higher computational cost.

## 4.2 Face Clustering

We consider the face clustering problem, where the goal is to group the face images into clusters according to their subjects. It is known that, under the Lambertian assumption, the face images of a subject with a fixed pose and varying illumination approximately lie in a linear subspace of dimension 9 [2]. So, a collection of face images of $k$ subjects approximately lie in a union of 9-dimensional subspaces. Therefore the face clustering problem can be solved by using subspace clustering methods.

We test on the Extended Yale B database [18]. This dataset consists of 2,414 frontal face images of 38 subjects under 9 poses and 64 illumination conditions. For each subject, there are 64 images. Each cropped face image consists of $192 \times 168$ pixels. To reduce the computation and memory cost, we downsample each image to $32 \times 32$ pixels and vectorize it to a 1,024 vector as a data point. Each data point is normalized to have a unit length. We then construct the data matrix $\mathbf{X}$ from subsets which consist of different numbers of subjects $k \in \{2, 3, 5, 8, 10\}$ from the Extended Yale B database. For each $k$, we randomly sample $k$ subjects face images from all 38 subjects to construct the data matrix $\mathbf{X} \in \mathbb{R}^{D \times n}$, where $D = 1024$ and $n = 64k$. Then the subspace clustering methods can be performed on $\mathbf{X}$ and the clustering error is recorded. We run 20 trials and the mean, median, and standard variance of clustering errors are reported.

TABLE 5: Clustering error (%) of different algorithms on the Extended Yale B database.

| method | 2 subjects | | | 3 subjects | | | 5 subjects | | | 8 subjects | | | 10 subjects | | |
|---|---|---|---|---|---|---|---|---|---|---|---|---|---|---|---|
| | mean | median | std | mean | median | std | mean | median | std | mean | median | std | mean | median | std |
| SCC | 24.02 | 19.92 | 17.82 | 42.19 | 41.93 | 8.93 | 61.36 | 62.34 | 6.10 | 71.87 | 72.27 | 4.72 | 72.48 | 73.28 | 6.14 |
| SSC | 1.64 | 0.78 | 2.91 | 3.26 | 0.52 | 7.69 | 6.30 | 4.22 | 5.43 | 8.94 | 9.67 | 6.18 | 10.09 | 11.33 | 4.59 |
| LRR | 5.39 | 0.39 | 14.50 | 6.04 | 1.04 | 12.34 | 8.13 | 2.34 | 9.61 | 6.79 | 3.42 | 6.50 | 9.49 | 12.58 | 5.38 |
| LSR | 3.16 | 0.78 | 10.18 | 3.96 | 1.56 | 8.72 | 7.85 | 6.72 | 8.72 | 28.14 | 31.05 | 12.32 | 33.27 | 33.12 | 4.57 |
| S³C | **1.29** | 0.00 | 2.69 | 2.79 | 0.52 | 7.38 | 4.66 | 1.88 | 5.15 | 6.37 | 6.35 | 5.32 | 6.87 | 6.17 | 3.67 |
| BDR-B | 3.28 | 0.78 | 10.15 | 3.02 | 1.30 | 7.78 | 4.45 | 2.19 | 6.29 | **3.08** | 2.93 | 1.18 | **2.95** | 2.81 | 1.09 |
| BDR-Z | 2.97 | 0.00 | 10.23 | **1.15** | 1.04 | 0.95 | **3.00** | 2.66 | 2.25 | 4.46 | 4.20 | 2.39 | 4.04 | 3.52 | 1.52 |

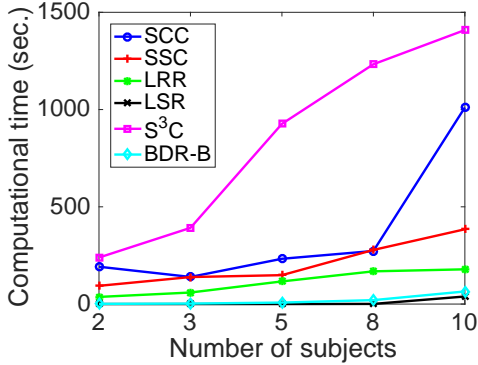

Fig. 4: Average computational time (sec.) of the algorithms on the Extended Yale B database as a function of the number of subjects.

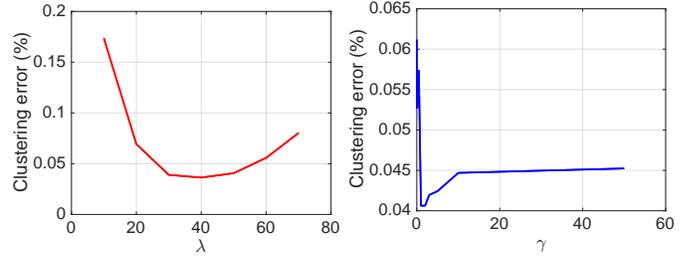

Fig. 5: Clustering error (%) of BDR-Z as a function of $\lambda$ when fixing $\gamma = 1$ (left) and $\gamma$ when fixing $\lambda = 50$ (right) for the 10 subjects problems from the Extended Yale B database.

The clustering errors by different subspace clustering methods on the Extended Yale B database are shown in Table 5. It can be seen that our BDR-B and BDR-Z achieve similar performance and both outperform other methods in most cases. Generally, when the number of subjects (or subspaces) increases, the clustering problem is more challenging. We find that the improvements by our methods are more significant when the number of subjects increases. This experiment clearly demonstrates the effectiveness of our BDR for the challenging face clustering task on the Extended Yale B database. S³C [19] is an improved SSC method and it also performs well in some cases. However, it needs to tune more parameters in order to achieve comparable performance and it is very time consuming. Figure 4 provides the average computational time of each method as a function of the number of subjects. It can be seen that S³C has the most highest computational time, while LSR, which has a closed form solution, is the most efficient method. Our BDR-B (BDR-Z has as similar running time and thus is not reported) is faster than most methods except LSR (LSR is much faster than BDR). So our BDR is the best choice when considering the trade-off between the performance and computational cost. Furthermore, we consider the influence of the parameters $\lambda$ and $\gamma$ on the clustering performance. On this dataset, we observe that $\lambda = 50$ and $\gamma = 1$ perform well in most cases. We report the average clustering error on the 10 subjects problem based on two settings: (1) fix $\gamma = 1$ and choose $\lambda \in \{10, 20, 30, 40, 50, 60, 70\}$; (2) fix $\lambda = 50$ and choose $\gamma \in \{0.001, 0.01, 0.1, 0.5, 1, 2, 3, 5, 10, 50\}$. The results are shown in Figure 5. It can be seen that the clustering error increases when $\lambda$ and $\gamma$ are relatively too small or too large. The reason for the performance degeneration in the "too small" case is that the regularization effect is relatively weak. On the other hand, if $\lambda$ and $\gamma$ are relatively large, $\mathbf{Z}$ and $\mathbf{B}$ in the early iterations are not discriminative due to relatively large loss $\|\mathbf{X} - \mathbf{X}\mathbf{Z}\|^2$. This issue may accumulate till the algorithm converges due to the nonconvexity of the problem and the non-optimal solution guarantee issue of our solver.

### 4.3 Handwritten Digit Clustering

We consider the application of subspace clustering for clustering images of handwritten digits which also have the subspace structure of dimension 12 [13]. We test on the MNIST database [17][4], which contains grey scale images of handwritten digits $0 \sim 9$. There are 10 subjects of digits. We consider the clustering problems with the number of subjects $k$ varying from 2 to 10. For each $k$, we run the experiments for 20 trials and report the mean clustering error. For each trial and each $k$, we consider random $k$ subjects of digits from $0 \sim 9$, and each subject has 100 randomly sampled images. Each grey image is of size $28 \times 28$ and is vectorized as a vector of length 784. Each data point is normalized to have a unit length. So for each $k$, we have the data matrix of size $784 \times 100k$.

---

4. We use the version at https://www.csie.ntu.edu.tw/~cjlin/libsvmtools/datasets/.

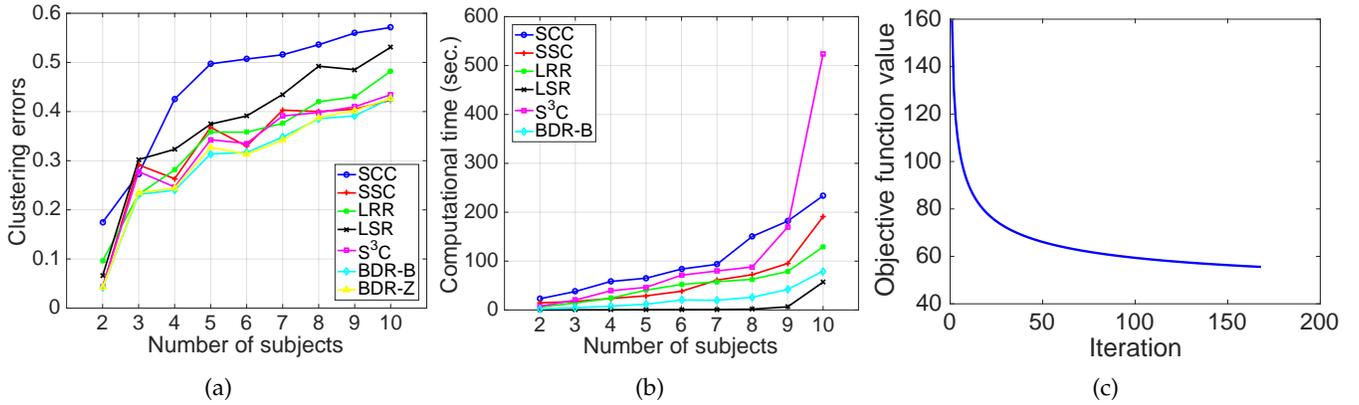

Fig. 6: Results on the MNIST database. (a) Plots of clustering errors v.s. the number of subjects; (b) Plots of average computational time (sec.) v.s. the number of subjects; (c) Plots of the objective function value of (14) v.s. iterations on a 5 subjects subset.

Figure 6 (a) plots the clustering errors as a function of the number of subjects on the MNIST database. It can be seen that our BDR-B and BDR-Z achieve the smallest clustering errors in most cases, though the improvements over the best compared method are different on different numbers of subjects. Figure 6 (b) gives a comparison on the average running time and it can be seen that our BDR-B (similar to BDR-Z) is much more efficient than most methods except LSR. The clustering performance of SSC and S³C is close to our BDR-B in some cases, but their computational cost is much higher than ours. So this experiment demonstrates the effectiveness and high-efficiency of our BDR. Furthermore, to verify our theoretical convergence results, we plot of the objective function value of (14) in each iteration obtained in Algorithm 1 for all iterations on a 5 subjects subset of the MNIST database in Figure 6 (c). It can be seen that the objective function value is monotonically decreasing and this phenomenon is consistent with our convergence analysis in Proposition 8.

## 5 CONCLUSION AND FUTURE WORKS

This paper studies the subspace clustering problem which aims to group the data points approximately drawn from a union of $k$ subspaces into $k$ clusters corresponding to their underlying subspaces. We observe that many existing spectral-type subspace clustering methods own the same block diagonal property under certain subspace assumption. We consider a general problem and show that if the objective satisfies the proposed Enforced Block Diagonal (EBD) conditions or its solution is unique, then the solution(s) obey the block diagonal property. This unified view provides insights into the relationship among the block diagonal property of the solution and the used objectives, as well as to facilitate the design of new algorithms. Inspired by the block diagonal property, we propose the first $k$-block diagonal regularizer which is useful for encouraging the matrix to be $k$-block diagonal. This leads to the Block Diagonal Representation (BDR) method for subspace clustering. A disadvantage of the BDR model is that it is nonconvex due to the $k$-block diagonal regularizer. We propose to solve the BDR model by a simple and generally efficient method and more importantly we provide the convergence guarantee without restrictive assumptions. Numerical experiments well demonstrate the effectiveness of our BDR.

There are many potential interesting future works:

1. The problem of the affinity matrix construction is not limited to the subspace clustering (or spectral clustering), but is everywhere and appears in many applications, e.g., [44], [46], [33]. The proposed $k$-block diagonal regularizer provides a new learning way and it is natural to consider the extension to related applications.

2. Beyond the sparse vector, low-rank matrix, the block diagonal matrix is another interesting structure of structured sparsity. The sparsity of the sparse vector is defined on the entries while the sparsity of the low-rank matrix is defined on the singular values. For the block diagonal matrix, its sparsity can be defined on the eigenvalues of the Laplacian matrix. So we can say that a block diagonal affinity matrix is *spectral sparse* if there have many connected blocks. This perspective motivates us to consider the statistical recovery guarantee of the block diagonal matrix regularized or constrained problems as that in compressive sensing.

3. The proposed $k$-block diagonal regularizer is nonconvex and this makes the optimization of the problem with such a regularizer challenging. Our proposed solver and convergence guarantee are specific for the nonconstrained BDR problem. How to solve the related problems but with a linear constraint and provide the convergence guarantee is interesting. The general Alternating Direction Method of Multipliers [24] is a potential solver.

## REFERENCES

[1] L. Bako. Identification of switched linear systems via sparse optimization. *Automatica*, 47(4):668–677, 2011.

[2] R. Basri and D. W. Jacobs. Lambertian reflectance and linear subspaces. *IEEE Trans. Pattern Recognition and Machine Intelligence*, 25(2):218–233, 2003.

[3] P. S. Bradley and O. L. Mangasarian. k-plane clustering. *Journal of Global Optimization*, 16(1):23–32, 2000.

[4] T. Brox and J. Malik. Object segmentation by long term analysis of point trajectories. In *Proc. European Conf. Computer Vision*, pages 282–295. Springer, 2010.

[5] E. J. Candès, X. Li, Y. Ma, and J. Wright. Robust principal component analysis? *J. ACM*, 58(3):11, 2011.

[6] E. J. Candès and B. Recht. Exact matrix completion via convex optimization. *Foundations of Computational Math.*, 9(6):717–772, 2009.

[7] G. Chen and G. Lerman. Spectral curvature clustering (SCC). *Int'l J. Computer Vision*, 81(3):317–330, 2009.

[8] J. P. Costeira and T. Kanade. A multibody factorization method for independently moving objects. *Int'l J. Computer Vision*, 29(3):159–179, 1998.

[9] J. Dattorro. Convex optimization & euclidean distance geometry. 2016. http://meboo.convexoptimization.com/Meboo.html.

[10] E. Elhamifar and R. Vidal. Sparse subspace clustering: Algorithm, theory, and applications. *IEEE Trans. Pattern Recognition and Machine Intelligence*, 35(11):2765–2781, 2013.

[11] J. Feng, Z. Lin, H. Xu, and S. Yan. Robust subspace segmentation with block-diagonal prior. In *Proc. IEEE Conf. Computer Vision and Pattern Recognition*, pages 3818–3825, 2014.

[12] C. W. Gear. Multibody grouping from motion images. *Int'l J. Computer Vision*, 29(2):133–150, 1998.

[13] T. Hastie and P. Y. Simard. Metrics and models for handwritten character recognition. *Statistical Science*, pages 54–65, 1998.

[14] H. Hu, Z. Lin, J. Feng, and J. Zhou. Smooth representation clustering. In *Proc. IEEE Conf. Computer Vision and Pattern Recognition*, pages 3834–3841, 2014.

[15] A. Jalali, Y. Chen, S. Sanghavi, and H. Xu. Clustering partially observed graphs via convex optimization. In *Proc. Int'l Conf. Machine Learning*, pages 1001–1008, 2011.

[16] K. Kanatani. Motion segmentation by subspace separation and model selection. In *Proc. IEEE Int'l Conf. Computer Vision*, pages 586–591, 2001.

[17] Y. LeCun, L. Bottou, Y. Bengio, and P. Haffner. Gradient-based learning applied to document recognition. *Proc. IEEE*, 86(11):2278–2324, 1998.

[18] K.-C. Lee, J. Ho, and D. J. Kriegman. Acquiring linear subspaces for face recognition under variable lighting. *IEEE Trans. Pattern Recognition and Machine Intelligence*, 27(5):684–698, 2005.

[19] C.-G. Li and R. Vidal. Structured sparse subspace clustering: a unified optimization framework. In *Proc. IEEE Conf. Computer Vision and Pattern Recognition*, pages 277–286, 2015.

[20] G. Liu, Z. Lin, S. Yan, J. Sun, Y. Yu, and Y. Ma. Robust recovery of subspace structures by low-rank representation. *IEEE Trans. Pattern Recognition and Machine Intelligence*, 35(1):171–184, 2013.

[21] G. Liu, Z. Lin, and Y. Yu. Robust subspace segmentation by low-rank representation. In *Proc. Int'l Conf. Machine Learning*, pages 663–670, 2010.

[22] G. Liu and S. Yan. Latent low-rank representation for subspace segmentation and feature extraction. In *Proc. IEEE Int'l Conf. Computer Vision*, pages 1615–1622, 2011.

[23] C. Lu, J. Feng, Z. Lin, and S. Yan. Correlation adaptive subspace segmentation by trace Lasso. In *Proc. IEEE Int'l Conf. Computer Vision*, pages 1345–1352, 2013.

[24] C. Lu, J. Feng, S. Yan, and Z. Lin. A unified alternating direction method of multipliers by majorization minimization. *arXiv:1607.02584*, 2016.

[25] C. Lu, J. Tang, M. Lin, L. Lin, S. Yan, and Z. Lin. Correntropy induced L2 graph for robust subspace clustering. In *Proc. IEEE Int'l Conf. Computer Vision*, 2013.

[26] C.-Y. Lu, H. Min, Z.-Q. Zhao, L. Zhu, D.-S. Huang, and S. Yan. Robust and efficient subspace segmentation via least squares regression. In *Proc. European Conf. Computer Vision*, 2012.

[27] D. Luo, F. Nie, C. Ding, and H. Huang. Multi-subspace representation and discovery. In *Joint European Conf. Machine Learning and Knowledge Discovery in Databases*, volume 6912 LNAI, pages 405–420, 2011.

[28] Y. Ma, A. Y. Yang, H. Derksen, and R. Fossum. Estimation of subspace arrangements with applications in modeling and segmenting mixed data. *SIAM Rev.*, 50.

[29] J. Mairal. Optimization with first-order surrogate functions. In *Proc. Int'l Conf. Machine Learning*, pages 783–791, 2013.

[30] B. Nasihatkon and R. Hartley. Graph connectivity in sparse subspace clustering. In *Proc. IEEE Conf. Computer Vision and Pattern Recognition*, pages 2137–2144, 2011.

[31] A. Y. Ng, M. I. Jordan, Y. Weiss, et al. On spectral clustering: Analysis and an algorithm. In *Advances in Neural Information Processing Systems*, pages 849–856, 2002.

[32] Y. Ni, J. Sun, X. Yuan, S. Yan, and L.-F. Cheong. Robust low-rank subspace segmentation with semidefinite guarantees. In *IEEE Int'l Conf. Data Mining Workshops*, pages 1179–1188, 2010.

[33] J. Shi and J. Malik. Normalized cuts and image segmentation. *IEEE Trans. Pattern Recognition and Machine Intelligence*, 22(8):888–905, 2000.

[34] M. Soltanolkotabi and E. J. Candès. A geometric analysis of subspace clustering with outliers. *The Annals of Statistics*, pages 2195–2238, 2012.

[35] M. Soltanolkotabi, E. Elhamifar, E. J. Candès, et al. Robust subspace clustering. *The Annals of Statistics*, 42(2):669–699, 2014.

[36] R. Tron and R. Vidal. A benchmark for the comparison of 3-D motion segmentation algorithms. In *Proc. IEEE Conf. Computer Vision and Pattern Recognition*, pages 1–8. IEEE, 2007.

[37] P. Tseng. Nearest $q$-flat to $m$ points. *J. Optimization Theory and Applications*, 105(1):249–252, 2000.

[38] R. Vidal, Y. Ma, and S. Sastry. Generalized principal component analysis (GPCA). *IEEE Trans. Pattern Recognition and Machine Intelligence*, 27(12):1945–1959, 2005.

[39] U. Von Luxburg. A tutorial on spectral clustering. *Statistics and Computing*, 17(4):395–416, 2007.

[40] S. Wang, X. Yuan, T. Yao, S. Yan, and J. Shen. Efficient subspace segmentation via quadratic programming. In *Proc. AAAI Conf. Artificial Intelligence*, volume 1, pages 519–524, 2011.

[41] Y.-X. Wang and H. Xu. Noisy sparse subspace clustering. In *Proc. Int'l Conf. Machine Learning*, pages 89–97, 2013.

[42] Y.-X. Wang, H. Xu, and C. Leng. Provable subspace clustering: When LRR meets SSC. In *Advances in Neural Information Processing Systems*, pages 64–72, 2013.

[43] J. Yan and M. Pollefeys. A general framework for motion segmentation: Independent, articulated, rigid, non-rigid, degenerate and non-degenerate. In *Proc. European Conf. Computer Vision*, pages 94–106. Springer, 2006.

[44] S. Yan, D. Xu, B. Zhang, H.-J. Zhang, Q. Yang, and S. Lin. Graph embedding and extensions: a general framework for dimensionality reduction. *IEEE Trans. Pattern Recognition and Machine Intelligence*, 29(1):40–51, 2007.

[45] A. Zhang, N. Fawaz, S. Ioannidis, and A. Montanari. Guess who rated this movie: Identifying users through subspace clustering. *arXiv:1208.1544*, 2012.

[46] X. Zhu and A. B. Goldberg. Introduction to semi-supervised learning. *Synthesis Lectures on Artificial Intelligence and Machine Learning*, 3(1):1–130, 2009.

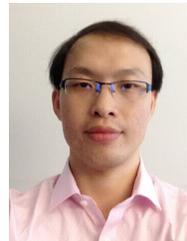

**Canyi Lu** is currently a Ph.D. student with the Department of Electrical and Computer Engineering at the National University of Singapore. His current research interests include computer vision, machine learning, pattern recognition and optimization. He was the winner of the Microsoft Research Asia Fellowship 2014.

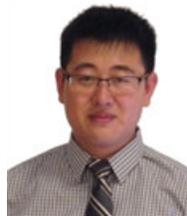

**Jiashi Feng** is currently an assistant Professor in the department of electrical and computer engineering in the National University of Singapore. He got his B.E. degree from University of Science and Technology, China in 2007 and Ph.D. degree from National University of Singapore in 2014. He was a postdoc researcher at University of California from 2014 to 2015. His current research interest focuses on machine learning and computer vision techniques for large-scale data analysis. Specifically, he has done work in object recognition, deep learning, machine learning, high-dimensional statistics and big data analysis.

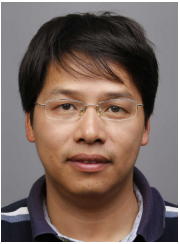

**Zhouchen Lin** (M00-SM08) received the Ph.D. degree in applied mathematics from Peking University in 2000. He is currently a Professor with the Key Laboratory of Machine Perception, School of Electronics Engineering and Computer Science, Peking University. He is also a chair professor of Northeast Normal University. His research areas include computer vision, image processing, machine learning, pattern recognition, and numerical optimization. He is an area chair of CVPR 2014/2016, ICCV 2015, and NIPS 2015, and a senior program committee member of AAAI 2016/2017 and IJCAI 2016. He is an Associate Editor of the IEEE Transactions on Pattern Analysis and Machine Intelligence and the International Journal of Computer Vision. He is an IAPR Fellow.

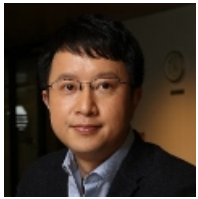

**Tao Mei** (M07-SM11) received the BE degree in automation and the PhD degree in pattern recognition and intelligent systems from the University of Science and Technology of China, Hefei, China, in 2001 and 2006, respectively. He is a lead researcher with Microsoft Research, Beijing, China. His current research interests include multimedia information retrieval and computer vision. He has authored or coauthored more than 150 papers in journals and conferences, 10 book chapters, and edited four books/proceedings. He holds 15 US granted patents and more than 20 in pending. He received several paper awards from prestigious multimedia journals and conferences, including the IEEE Circuits and Systems Society Circuits and Systems for Video Technology Best Paper Award in 2014, the IEEE Transactions on Multimedia Prize Paper Award in 2013, the Best Student Paper Award at the IEEE VCIP in 2012, and the Best Paper Awards at ACM Multimedia in 2009 and 2007, etc. He is on the editorial board of IEEE Transactions on Multimedia, ACM Transactions on Multimedia Computing, Communications, and Applications, Machine Vision and Applications, and Multimedia Systems, and was an associate editor of Neurocomputing, a guest editor of seven international journals. He is the general cochair of ACM ICIMCS 2013, the program co-chair of ACM Multimedia 2018, IEEE ICME 2015, IEEE MMSP 2015, and MMM 2013, and the area chair for a dozen conferences. He is a senior member of the IEEE and the ACM.

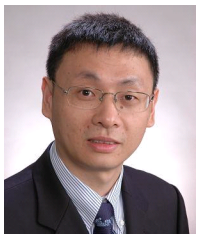

**Shuicheng Yan** is currently an Associate Professor at the Department of Electrical and Computer Engineering at National University of Singapore, and the founding lead of the Learning and Vision Research Group (http://www.lv-nus.org). Dr. Yan's research areas include machine learning, computer vision and multimedia, and he has authored/co-authored hundreds of technical papers over a wide range of research topics, with Google Scholar citation >31,000 times and H-index 67. He is ISI Highly-cited Researcher, 2014 and IAPR Fellow 2014. He has been serving as an associate editor of IEEE TKDE, TCSVT and ACM Transactions on Intelligent Systems and Technology (ACM TIST). He received the Best Paper Awards from ACM MM'13 (Best Paper and Best Student Paper), ACM MM12 (Best Demo), PCM'11, ACM MM10, ICME10 and ICIMCS'09, the runner-up prize of ILSVRC'13, the winner prize of ILSVRC14 detection task, the winner prizes of the classification task in PASCAL VOC 2010-2012, the winner prize of the segmentation task in PASCAL VOC 2012, the honourable mention prize of the detection task in PASCAL VOC'10, 2010 TCSVT Best Associate Editor (BAE) Award, 2010 Young Faculty Research Award, 2011 Singapore Young Scientist Award, and 2012 NUS Young Researcher Award.

# Supplementary Material

This document gives the proofs of some propositions and theorems. We continue the number of equations, propositions and theorems in the main submission of the paper. Please read the main submission together with this document.

## APPENDIX A
## PROOFS OF THE BLOCK DIAGONAL PROPERTY

### A.1 Proof of Proposition 1

*Proof.* The result is obvious by using the definitions of EBD conditions. □

### A.2 Proof of Proposition 2

*Proof.* We are given the function $f(\mathbf{Z}, \mathbf{X}) = \sum_{ij} g_{ij}(z_{ij})$ which is separable w.r.t. each entry $z_{ij}$ of $\mathbf{Z}$. First, for any permutation matrix $\mathbf{P}$, $\mathbf{P}^\top \mathbf{Z} \mathbf{P}$ keeps the same entries in $\mathbf{Z}$ but rearranges their positions. Both $\mathbf{P}^\top \mathbf{Z} \mathbf{P}$ and $\mathbf{Z}$ have the same entries. So the separability of $f$ guarantees that the EBD condition (1) holds. The EBD condition (2) also holds since it is equivalent to the given assumptions that $g_{ij}(z_{ij}) \geq 0$, $g_{ij}(z_{ij}) = 0$ if and only if $z_{ij} = 0$ for all $i$ and j. The EBD condition (3) naturally holds due to the separability of $f$. □

### A.3 Proof of Proposition 3

*Proof.* We are given the function $f(\mathbf{Z}, \mathbf{X}) = \sum_j g_j([\mathbf{Z}]_{:,j}, \mathbf{X})$ which is separable w.r.t. each column $[\mathbf{Z}]_{:,j}$ of $\mathbf{Z}$. First, we verify the EBD condition (1). For any permutation matrix $\mathbf{P}$, we have

$$f(\mathbf{P}^\top \mathbf{Z} \mathbf{P}, \mathbf{X} \mathbf{P}) = \sum_j g_j([\mathbf{P}^\top \mathbf{Z} \mathbf{P}]_{:,j}, \mathbf{X} \mathbf{P})$$

$$= \sum_j g_j(\mathbf{P}^\top [\mathbf{Z} \mathbf{P}]_{:,j}, \mathbf{X} \mathbf{P}) \tag{22}$$

$$= \sum_j g_j(\mathbf{P}^\top [\mathbf{Z}]_{:,j}, \mathbf{X} \mathbf{P}) \tag{23}$$

$$= \sum_j g_j([\mathbf{Z}]_{:,j}, \mathbf{X}) \tag{24}$$

$$= f(\mathbf{Z}, \mathbf{X}),$$

where (22) uses the simple property $[\mathbf{A} \mathbf{B}]_{:,j} = \mathbf{A} [\mathbf{B}]_{:,j}$, (23) uses that fact that $\mathbf{A} \mathbf{P}$ reorders the columns of the matrix $\mathbf{A}$ and keeps the column entries, and (24) uses the given assumption $g_j(\mathbf{w}, \mathbf{X}) = g_j(\mathbf{P}^\top \mathbf{w}, \mathbf{X} \mathbf{P})$. This means that the EBD condition (1) holds.

Second, to verify the EBD condition (2), we deduce

$$f(\mathbf{Z}, \mathbf{X}) = \sum_j g_j([\mathbf{Z}]_{:,j}, \mathbf{X}) \geq \sum_j g_j([\mathbf{Z}^B]_{:,j}, \mathbf{X}) = f(\mathbf{Z}^B, \mathbf{X}), \tag{25}$$

where we use the given assumption $g_j(\mathbf{w}, \mathbf{X}) \geq g_j(\mathbf{w}^B, \mathbf{X})$. Note that it is further assumed that the equality holds if and only if $\mathbf{w} = \mathbf{w}^B$. This is implies that the equality in (25) holds if and only if $\mathbf{Z} = \mathbf{Z}^B$. Thus, the EBD condition (2) holds.

Third, to verify the EBD condition (3), we deduce

$$f(\mathbf{Z}^B, \mathbf{X}) = \sum_j g_j([\mathbf{Z}^B]_{:,j}, \mathbf{X})$$

$$= \sum_k g_k \left( \begin{bmatrix} \mathbf{Z}_1 \\ \mathbf{0} \end{bmatrix}_{:,k}, \mathbf{X} \right) + \sum_l g_l \left( \begin{bmatrix} \mathbf{0} \\ \mathbf{Z}_2 \end{bmatrix}_{:,l}, \mathbf{X} \right) \tag{26}$$

$$= \sum_k g_k([\mathbf{Z}_1]_{:,k}, \mathbf{X}_1) + \sum_l g_l([\mathbf{Z}_2]_{:,l}, \mathbf{X}_2) \tag{27}$$

$$= f(\mathbf{Z}_1, \mathbf{X}_1) + f(\mathbf{Z}_2, \mathbf{X}_2),$$

where (26) uses the definition of $\mathbf{Z}^B$ in the EBD conditions, and (27) uses the given assumption $g_j(\mathbf{w}^B, \mathbf{X}) = g_j(\mathbf{w}_1, \mathbf{X}_1)$. Thus, the EBD condition (3) holds. □

### A.4 Proof of Proposition 4

*Proof.* We are given the function $f(\mathbf{Z}, \mathbf{X}) = \sum_i g_i([\mathbf{Z}]_{i,:}, \mathbf{X})$ which is separable w.r.t. each row $[\mathbf{Z}]_{i,:}$ of $\mathbf{Z}$. First, we verify the EBD condition (1). For any permutation matrix $\mathbf{P}$, we have

$$f(\mathbf{P}^\top \mathbf{Z} \mathbf{P}, \mathbf{X} \mathbf{P}) = \sum_i g_i([\mathbf{P}^\top \mathbf{Z} \mathbf{P}]_{i,:}, \mathbf{X} \mathbf{P})$$

$$= \sum_i g_i([\mathbf{Z} \mathbf{P}]_{i,:}, \mathbf{X} \mathbf{P}) \tag{28}$$

$$= \sum_i g_i(([\mathbf{Z}]_{i,:})\mathbf{P}, \mathbf{X} \mathbf{P}) \tag{29}$$

$$= \sum_i g_i([\mathbf{Z}]_{i,:}, \mathbf{X}) \tag{30}$$

$$= f(\mathbf{Z}, \mathbf{X}),$$

where (28) uses the fact that $\mathbf{P}^\top \mathbf{A}$ reorders the rows of the matrix $\mathbf{A}$ and keeps the row entries, (29) uses the simple property $[\mathbf{A}\mathbf{B}]_{i,:} = [\mathbf{A}_{i,:}]\mathbf{B}$, and (30) uses the given assumption $g_i(\mathbf{w}^\top, \mathbf{X}) = g_i(\mathbf{w}^\top \mathbf{P}, \mathbf{X} \mathbf{P})$. This means that the EBD condition (1) holds.

Second, to verify the EBD condition (2), we deduce

$$f(\mathbf{Z}, \mathbf{X}) = \sum_i g_i([\mathbf{Z}]_{i,:}, \mathbf{X}) \geq \sum_i g_i([\mathbf{Z}^B]_{i,:}, \mathbf{X}) = f(\mathbf{Z}^B, \mathbf{X}), \tag{31}$$

where we use the given assumption $g_i(\mathbf{w}^\top, \mathbf{X}) \geq g_i((\mathbf{w}^B)^\top, \mathbf{X})$. Note that it is further assumed that the equality holds if and only if $\mathbf{w} = \mathbf{w}^B$. This is implies that the equality in (31) holds if and only if $\mathbf{Z} = \mathbf{Z}^B$. Thus, the EBD condition (2) holds.

Third, to verify the EBD condition (3), we deduce

$$f(\mathbf{Z}^B, \mathbf{X}) = \sum_i g_i([\mathbf{Z}^B]_{i,:}, \mathbf{X})$$

$$= \sum_k g_k([\mathbf{Z}_1, 0]_{k,:}, \mathbf{X}) + \sum_l g_l([0, \mathbf{Z}_2]_{l,:}, \mathbf{X}) \tag{32}$$

$$= \sum_k g_k([\mathbf{Z}_1]_{k,:}, \mathbf{X}_1) + \sum_l g_l([\mathbf{Z}_2]_{l,:}, \mathbf{X}_2) \tag{33}$$

$$= f(\mathbf{Z}_1, \mathbf{X}_1) + f(\mathbf{Z}_2, \mathbf{X}_2),$$

where (32) uses the definition of $\mathbf{Z}^B$ in the EBD conditions, and (33) uses the given assumption $g_i((\mathbf{w}^B)^\top, \mathbf{X}) = g_i(\mathbf{w}_1^\top, \mathbf{X}_1)$. Thus, the EBD condition (3) holds. $\qquad \square$

### A.5 Proof of Proposition 5

*Proof.* We are given a series of functions $f_i$'s which satisfy the EBD conditions (1)-(3). It is easy to verify that their positive combination, i.e., $\lambda_i f_i$, where $\lambda_i > 0$, still satisfies the EBD conditions (1)-(3) by directly using the definitions of EBD conditions. $\qquad \square$

### A.6 Proof of Proposition 6

*Proof.* We are given a function $f_1$ which satisfies the EBD conditions (1)-(3) and a function $f_2$ which satisfies the EBD conditions (1)(3) and the first part of EBD condition (2). It is obvious that the EBD conditions (1)(3) and the first part of (2) holds for $f = f_1 + f_2$ by directly using the definitions of EBD conditions. For the second part of the EBD condition (2), it is easy to see that it still holds for $f$ since $f_1$ satisfies the EBD condition (2) and $f_2$ satisfies the first part of EBD condition (2). $\qquad \square$

### A.7 Proof of Theorem 2

*Proof.* First, $\|\mathbf{Z}\|_0$, $\|\mathbf{Z}\|_1$, $\|\mathbf{Z}\|^2$, $\|\mathbf{Z}\|_1 + \lambda \|\mathbf{Z}\|^2$, and $\sum_{ij} \lambda_{ij} |z_{ij}|^{p_{ij}}$ are separable w.r.t. $z_{ij}$'s. By Proposition 2, the EBD conditions (1)-(3) hold for these functions. The EBD conditions for $\|\mathbf{Z}\|_1 + \lambda \|\mathbf{Z}\|^2$ can also be verified by using Proposition 5.

Second, we prove that the EBD conditions hold for the $\ell_{2,1}$-norm $\|\mathbf{Z}\|_{2,1}$ and trace Lasso $\sum_j \|\mathbf{X} \text{Diag}([\mathbf{Z}]_{:,j})\|_*$ by using Proposition 3. Consider $\|\mathbf{Z}\|_{2,1} = \sum_j g([\mathbf{Z}]_{:,j})$, where $g$ is the $\ell_2$-norm. It is obvious that the $\ell_2$-norm satisfies all the three conditions in Proposition 3, and thus the result holds for $\|\mathbf{Z}\|_{2,1}$. Consider $\sum_j \|\mathbf{X} \text{Diag}([\mathbf{Z}]_{:,j})\|_* = \sum_j g([\mathbf{Z}]_{:,j}, \mathbf{X})$, where $g(\mathbf{w}, \mathbf{X}) = \|\mathbf{X}\mathrm{Diag}(\mathbf{w})\|_*$ is the trace Lasso. Now, we verify the conditions in Proposition 3 for trace Lasso. For the first condition, consider any permutation matrix $\mathbf{P}$ of proper size, we have

$$g(\mathbf{P}^\top \mathbf{w}, \mathbf{X}\mathbf{P}) = \left\|\mathbf{X}\mathbf{P}\mathrm{Diag}(\mathbf{P}^\top \mathbf{w})\right\|_* = \left\|\mathbf{X}\mathbf{P}\mathbf{P}^\top\mathrm{Diag}(\mathbf{w})\right\|_* = \|\mathbf{X}\mathrm{Diag}(\mathbf{w})\|_* = g(\mathbf{w}, \mathbf{X}),$$

where we use the fact that the permutation matrix is orthogonal. For the second condition, we partition $\mathbf{X} = [\mathbf{X}_1, \mathbf{X}_2]$ according to $\mathbf{w} = [\mathbf{w}_1; \mathbf{w}_2]$. We have

$$g(\mathbf{w}, \mathbf{X}) = \|\mathbf{X}\mathrm{Diag}(\mathbf{w})\|_* = \|[\mathbf{X}_1\mathrm{Diag}(\mathbf{w}_1)\ \mathbf{X}_2\mathrm{Diag}(\mathbf{w}_2)]\|_* \geq \|[\mathbf{X}_1\mathrm{Diag}(\mathbf{w}_1)\ 0]\|_* = \left\|\mathbf{X}\mathrm{Diag}(\mathbf{w}^B)\right\|_* = g(\mathbf{w}^B, \mathbf{X}), \quad (34)$$

where the inequality is obtained by using Lemma 11 in [32] and note that the equality holds if and only if $\mathbf{X}_2\mathrm{Diag}(\mathbf{w}_2) = 0$. This is equivalent to $\mathbf{w}_2 = 0$ or $\mathbf{w} = \mathbf{w}^B$ since $[\mathbf{X}]_{:,j} \neq 0$ for all $j$. The third condition in Proposition 3 is given by the last second equation of (34). Thus, by Proposition 3, the EBD conditions hold for $\sum_j \|\mathbf{X}\mathrm{Diag}([\mathbf{z}]_{:,j})\|_*$.

Third, the $\ell_{1,2}$-norm $\|\mathbf{Z}\|_{1,2}$ is row separable. It is easy to verify that it satisfies all the three conditions in Proposition 4 and thus it satisfies the EBD conditions.

Fourth, we show that the EBD conditions hold for the $\ell_1$+nuclear norm $\|\mathbf{Z}\|_1 + \lambda \|\mathbf{Z}\|_*$ by using Proposition 6. We know that $\|\mathbf{Z}\|_1$ satisfies the EBD conditions (1)-(3). For $\|\mathbf{Z}\|_*$, it is obvious that $\|\mathbf{Z}\|_* = \left\|\mathbf{P}^\top \mathbf{Z}\mathbf{P}\right\|_*$ for any permutation matrix $\mathbf{P}$ which is orthogonal. Also, by Lemma 7.4 in [21], we have

$$\left\|\begin{bmatrix}\mathbf{Z}_1 & \mathbf{Z}_3 \\ \mathbf{Z}_4 & \mathbf{Z}_2\end{bmatrix}\right\|_* \geq \left\|\begin{bmatrix}\mathbf{Z}_1 & 0 \\ 0 & \mathbf{Z}_2\end{bmatrix}\right\|_* = \|\mathbf{Z}_1\|_* + \|\mathbf{Z}_2\|_*.$$

Thus, the EBD conditions (1)(3) and the first part of EBD condition (2) hold for $\|\mathbf{Z}\|_*$. Hence, the three EBD conditions hold for $\|\mathbf{Z}\|_1 + \lambda \|\mathbf{Z}\|_*$ with $\lambda > 0$ by using Proposition 6.

At last, we show that the EBD conditions hold for $\|\mathbf{Z}^\top \mathbf{Z}\|_1$ when $\mathbf{Z} \geq 0$. For any permutation matrix $\mathbf{P}$, we have

$$\left\|(\mathbf{P}^\top \mathbf{Z}\mathbf{P})^\top (\mathbf{P}^\top \mathbf{Z}\mathbf{P})\right\|_1 = \left\|\mathbf{P}^\top \mathbf{Z}\mathbf{P}\mathbf{P}^\top \mathbf{Z}\mathbf{P}\right\|_1 = \left\|\mathbf{P}^\top \mathbf{Z}^\top \mathbf{Z}\mathbf{P}\right\|_1 = \left\|\mathbf{Z}^\top \mathbf{Z}\right\|_1,$$

where the last equation uses the fact $\mathbf{P}^\top \mathbf{Z}^\top \mathbf{Z}\mathbf{P}$ has the same entries as $\mathbf{Z}^\top \mathbf{Z}$, though the positions are different. Thus, the EBD condition (1) holds. For EBD condition (2), we decompose $\mathbf{Z} = \mathbf{Z}^B + \mathbf{Z}^C$, where

$$\mathbf{Z} = \begin{bmatrix}\mathbf{Z}_1 & \mathbf{Z}_3 \\ \mathbf{Z}_4 & \mathbf{Z}_2\end{bmatrix}, \ \mathbf{Z}^B = \begin{bmatrix}\mathbf{Z}_1 & 0 \\ 0 & \mathbf{Z}_2\end{bmatrix}, \ \mathbf{Z}^C = \begin{bmatrix}0 & \mathbf{Z}_3 \\ \mathbf{Z}_4 & 0\end{bmatrix}.$$

Then we have

$$\begin{aligned}
\left\|\mathbf{Z}^\top \mathbf{Z}\right\|_1 &= \left\|(\mathbf{Z}^B + \mathbf{Z}^C)^\top (\mathbf{Z}^B + \mathbf{Z}^C)\right\|_1 \\
&= \left\|(\mathbf{Z}^B)^\top \mathbf{Z}^B + (\mathbf{Z}^C)^\top \mathbf{Z}^C + (\mathbf{Z}^B)^\top \mathbf{Z}^C + (\mathbf{Z}^C)^\top \mathbf{Z}^B\right\|_1 \\
&\geq \left\|(\mathbf{Z}^B)^\top \mathbf{Z}^B\right\|_1,
\end{aligned}$$

where the inequality uses $\mathbf{Z} \geq 0$. Also, the inequality holds if and only if $\mathbf{Z}^C = 0$ or $\mathbf{Z} = \mathbf{Z}^B$. Thus, the EBD condition (2) holds. The EBD condition (3) also holds since

$$\left\|(\mathbf{Z}^B)^\top \mathbf{Z}^B\right\|_1 = \left\|(\mathbf{Z}_1)^\top \mathbf{Z}_1\right\|_1 + \left\|(\mathbf{Z}_2)^\top \mathbf{Z}_2\right\|_1.$$

The proof is completed. □

# APPENDIX B
## PROOFS OF THE CONVERGENCE OF ALGORITHM 1
### B.1 Proof of Proposition 7

*Proof.* It is obvious that problem (20) is equivalent to

$$\min_{\mathbf{B}} \ \frac{1}{2}\|\mathbf{B} - \hat{\mathbf{A}}\|^2, \ \text{s.t. } \mathbf{B} \geq 0, \mathbf{B} = \mathbf{B}^\top. \quad (35)$$

The constraint $\mathbf{B} = \mathbf{B}^\top$ suggests that $\|\mathbf{B} - \hat{\mathbf{A}}\|^2 = \|\mathbf{B} - \hat{\mathbf{A}}^\top\|^2$. Thus

$$\frac{1}{2}\|\mathbf{B} - \hat{\mathbf{A}}\|^2 = \frac{1}{4}\|\mathbf{B} - \hat{\mathbf{A}}\|^2 + \frac{1}{4}\|\mathbf{B} - \hat{\mathbf{A}}^\top\|^2 = \frac{1}{2}\left\|\mathbf{B} - (\hat{\mathbf{A}} + \hat{\mathbf{A}}^\top)/2\right\|^2 + c(\hat{\mathbf{A}}),$$

where $c(\hat{\mathbf{A}})$ only depends on $\hat{\mathbf{A}}$. Hence (35) is equivalent to

$$\min_{\mathbf{B}} \ \frac{1}{2}\left\|\mathbf{B} - (\hat{\mathbf{A}} + \hat{\mathbf{A}}^\top)/2\right\|^2, \ \text{s.t. } \mathbf{B} \geq 0, \mathbf{B} = \mathbf{B}^\top,$$

which has the solution $\mathbf{B}^* = \left[(\hat{\mathbf{A}} + \hat{\mathbf{A}}^\top)/2\right]_+$. □

### B.2 Proof of Proposition 8

*Proof.* First, from the optimality of $\mathbf{W}^{k+1}$ to (15), we have

$$f(\mathbf{Z}^k, \mathbf{B}^k, \mathbf{W}^{k+1}) + \iota_{S_2}(\mathbf{W}^{k+1}) \leq f(\mathbf{Z}^k, \mathbf{B}^k, \mathbf{W}^k) + \iota_{S_2}(\mathbf{W}^k). \tag{36}$$

Second, from the updating rule of $\mathbf{Z}^{k+1}$ in (16), we have

$$\mathbf{Z}^{k+1} = \arg\min_{\mathbf{Z}} \ f(\mathbf{Z}, \mathbf{B}^k, \mathbf{W}^{k+1}).$$

Note that $f(\mathbf{Z}, \mathbf{B}^k, \mathbf{W}^{k+1})$ is $\lambda$-strongly convex. We have

$$f(\mathbf{Z}^{k+1}, \mathbf{B}^k, \mathbf{W}^{k+1}) \leq f(\mathbf{Z}^k, \mathbf{B}^k, \mathbf{W}^{k+1}) - \frac{\lambda}{2} \left\| \mathbf{Z}^{k+1} - \mathbf{Z}^k \right\|^2, \tag{37}$$

where we use the Lemma B.5 in [29]. Third, from the updating rule of $\mathbf{B}^{k+1}$ in (19), we have

$$\mathbf{B}^{k+1} = \arg\min_{\mathbf{B}} \ f(\mathbf{Z}^{k+1}, \mathbf{B}, \mathbf{W}^{k+1}) + \iota_{S_1}(\mathbf{B}).$$

Note that $f(\mathbf{Z}^{k+1}, \mathbf{B}, \mathbf{W}^{k+1}) + \iota_{S_1}(\mathbf{B})$ is $\lambda$-strongly convex w.r.t. $\mathbf{B}$. We have

$$f(\mathbf{Z}^{k+1}, \mathbf{B}^{k+1}, \mathbf{W}^{k+1}) + \iota_{S_1}(\mathbf{B}^{k+1}) \leq f(\mathbf{Z}^{k+1}, \mathbf{B}^k, \mathbf{W}^{k+1}) + \iota_{S_1}(\mathbf{B}^k) - \frac{\lambda}{2} \left\| \mathbf{B}^{k+1} - \mathbf{B}^k \right\|^2.$$

Combining (36)-(38), we have

$$\begin{aligned} &f(\mathbf{Z}^{k+1}, \mathbf{B}^{k+1}, \mathbf{W}^{k+1}) + \iota_{S_1}(\mathbf{B}^{k+1}) + \iota_{S_2}(\mathbf{W}^{k+1}) \\ \leq &f(\mathbf{Z}^k, \mathbf{B}^k, \mathbf{W}^k) + \iota_{S_1}(\mathbf{B}^k) + \iota_{S_2}(\mathbf{W}^k) - \frac{\lambda}{2} \left\| \mathbf{B}^{k+1} - \mathbf{B}^k \right\|^2 - \frac{\lambda}{2} \left\| \mathbf{Z}^{k+1} - \mathbf{Z}^k \right\|^2. \end{aligned} \tag{38}$$

Hence, $f(\mathbf{Z}^k, \mathbf{B}^k, \mathbf{W}^k) + \iota_{S_1}(\mathbf{B}^k) + \iota_{S_2}(\mathbf{W}^k)$ is monotonically decreasing and thus it is upper bounded. This implies that $\{\mathbf{Z}^k\}$ and $\{\mathbf{B}^k\}$ are bounded. Also, $\mathbf{W}^k \in S_2$ implies that $\|\mathbf{W}^k\|_2 \leq 1$ and thus $\{\mathbf{W}^k\}$ is bounded.

Note that $\mathbf{W}^k$ and $\text{Diag}(\mathbf{B}^k \mathbf{1}) - \mathbf{B}^k$ are positive semi-definite. We have $\langle \text{Diag}(\mathbf{B}^k \mathbf{1}) - \mathbf{B}^k, \mathbf{W}^k \rangle \geq 0$. Thus $f(\mathbf{Z}^k, \mathbf{B}^k, \mathbf{W}^k) + \iota_{S_1}(\mathbf{B}^k) + \iota_{S_2}(\mathbf{W}^k) \geq 0$. Now, summing (38) over $k = 0, 1, \cdots$ we have

$$\sum_{k=0}^{+\infty} \frac{\lambda}{2} \left( \left\| \mathbf{B}^{k+1} - \mathbf{B}^k \right\|^2 + \left\| \mathbf{Z}^{k+1} - \mathbf{Z}^k \right\|^2 \right) \leq f(\mathbf{Z}^0, \mathbf{B}^0, \mathbf{W}^0).$$

This implies

$$\mathbf{B}^{k+1} - \mathbf{B}^k \to 0, \tag{39}$$

and

$$\mathbf{Z}^{k+1} - \mathbf{Z}^k \to 0. \tag{40}$$

By (39) and the updating of $\mathbf{W}^{k+1}$ in (15), we have

$$\mathbf{W}^{k+1} - \mathbf{W}^k \to 0. \tag{41}$$

The proof is completed. $\qquad \square$

### B.3 Proof of Theorem 6

*Proof.* Now, from the boundedness of $\{\mathbf{Z}^k, \mathbf{B}^k, \mathbf{W}^k\}$, there exists a point $(\mathbf{Z}^*, \mathbf{B}^*, \mathbf{W}^*)$ and a subsequence $\{\mathbf{Z}^{k_j}, \mathbf{B}^{k_j}, \mathbf{W}^{k_j}\}$ such that $\mathbf{Z}^{k_j} \to \mathbf{Z}^*$, $\mathbf{B}^{k_j} \to \mathbf{B}^*$, and $\mathbf{W}^{k_j} \to \mathbf{W}^*$. Then by (39)-(41), we have $\mathbf{Z}^{k_j+1} \to \mathbf{Z}^*$, $\mathbf{B}^{k_j+1} \to \mathbf{B}^*$ and $\mathbf{W}^{k_j+1} \to \mathbf{W}^*$. On the other hand, from the optimality of $\mathbf{W}^{k_j+1}$ to (15), $\mathbf{Z}^{k_j+1}$ to (16) and $\mathbf{B}^{k_j+1}$ to (17), we have

$$0 \in \nabla f_{\mathbf{W}}(\mathbf{Z}^{k_j}, \mathbf{B}^{k_j}, \mathbf{W}^{k_j+1}) + \partial_{\mathbf{W}} \iota_{S_2}(\mathbf{W}^{k_j+1}), \tag{42}$$

$$0 \in \nabla f_{\mathbf{Z}}(\mathbf{Z}^{k_j+1}, \mathbf{B}^{k_j}, \mathbf{W}^{k_j+1}), \tag{43}$$

$$0 \in \nabla f_{\mathbf{B}}(\mathbf{Z}^{k_j+1}, \mathbf{B}^{k_j+1}, \mathbf{W}^{k_j+1}) + \partial_{\mathbf{B}} \iota_{S_1}(\mathbf{B}^{k_j+1}). \tag{44}$$

Let $k \to +\infty$ in (42)-(44). We have

$$0 \in \nabla f_{\mathbf{W}}(\mathbf{Z}^*, \mathbf{B}^*, \mathbf{W}^*) + \partial_{\mathbf{W}} \iota_{S_2}(\mathbf{W}^*),$$
$$0 \in \nabla f_{\mathbf{Z}}(\mathbf{Z}^*, \mathbf{B}^*, \mathbf{W}^*),$$
$$0 \in \nabla f_{\mathbf{B}}(\mathbf{Z}^*, \mathbf{B}^*, \mathbf{W}^*) + \partial_{\mathbf{B}} \iota_{S_1}(\mathbf{B}^*).$$

Thus $(\mathbf{Z}^*, \mathbf{B}^*, \mathbf{W}^*)$ is a stationary point of (14). $\qquad \square$